\documentclass{article}

\PassOptionsToPackage{numbers, compress}{natbib}
\usepackage[preprint]{neurips_2026}

\usepackage{amssymb}
\usepackage[utf8]{inputenc} % allow utf-8 input
\usepackage[T1]{fontenc}    % use 8-bit T1 fonts
\usepackage{hyperref}       % hyperlinks
\usepackage{url}            % simple URL typesetting
\usepackage{booktabs}       % professional-quality tables
\usepackage{amsfonts}       % blackboard math symbols
\usepackage{nicefrac}       % compact symbols for 1/2, etc.
\usepackage{microtype}      % microtypography
\usepackage[table]{xcolor}         % colors
\usepackage{amsmath}
\usepackage{xspace}
\usepackage{cleveref}
\usepackage{multirow}
\usepackage{enumitem}

\usepackage{adjustbox}
\usepackage{makecell}
\usepackage{wrapfig}
\usepackage{tabularx}
\usepackage{siunitx}
\usepackage{caption}
\usepackage{placeins}
\graphicspath{{images/}{images/illustrative_images/}}

\newcommand{\algo}{LLaDA-Guard\xspace}

\newcommand{\circled}[1]{\raisebox{0.14em}{\textcircled{\scalebox{0.7}{#1}}}}

\title{Safety Reconstructed: Generative Modeling via
Masked Diffusion Builds Strong Safety Guardrails}
\author{%
  \textbf{Gert Lek}$^{1}$ \quad
  \textbf{Abele M\u{a}lan}$^{1}$ \quad
  \textbf{Chaoyi Zhu}$^{2}$ \\[3pt]
  \textbf{Pin-Yu Chen}$^{3}$ \quad
  \textbf{Robert Birke}$^{4}$ \quad
  \textbf{Lydia Chen}$^{1,2}$ \\[6pt]
  \normalfont
  $^{1}$University of Neuch\^atel \quad
  $^{2}$Delft University of Technology \\
  \normalfont
  $^{3}$IBM Research \quad
  $^{4}$University of Turin \\[3pt]
  \normalfont\texttt{gert.lek@unine.ch}
}

\begin{document}

\maketitle
\vspace{-10pt}

\begin{abstract}
Guard models are the last line of defense between a language model and a harmful output, yet their training objective is surprisingly narrow.
Existing guards learn to predict a single verdict token from a conversational context, concentrating supervision on a single target. The consequences are structural: models latch onto shortcut features, are overconfident, and remain sensitive to where safety evidence appears in the sequence rather than its role in the full context.
We propose a different framing. Rather than predicting a label from text, our \textbf{LLaDA-Guard} asks which label better explains the text: scoring the prompt or response under each label hypothesis and classifying based on their difference. This shifts supervision to every token in the moderated region, forcing the model to account for full content rather than its most discriminative fragments. We instantiate this idea with a masked diffusion language model, fine-tuning LLaDA-8B-Instruct with a class-conditional reconstruction objective using LoRA and requiring no architectural changes beyond the base model.
\algo leads on average rank against discriminative baselines trained on stronger backbones across seven held-out safety benchmarks,
while exhibiting substantially better confidence calibration (ECE 0.0875 vs. 0.1384 for Qwen3Guard), less over-defense on benign prompts with \texttt{unsafe}-looking cues, and less prompt leakage when moderating responses. Its generative nature further enables token-level risk localization as a natural byproduct, yielding a pipeline for rewriting \texttt{unsafe} prompts into \texttt{safe} equivalents without additional training and achieving a $60.7\%$ average conversion-to-\texttt{safe} rate.
\end{abstract}

\section{Introduction}
Large language models (LLMs) are increasingly deployed across consumer assistants, developer tools, and high-stakes domains such as healthcare and education \citep{llm-software-engineering,llm-education,llm-assistant-healthcare}. At this scale, a non-trivial share of interactions involves inappropriate or malicious requests~\citep{weidinger2021ethical, hendrycks2023overview, anwar2024foundational}, and even well-aligned models remain susceptible to jailbreaks, which reliably bypass their built-in refusal behavior~\citep{wei2023jailbroken, zou2023universal, shen2024anythingnow}. The long tail of usage uncovers failure modes not anticipated during alignment training~\citep{zheng2024lmsys, zhao2024wildchat}.

Content moderation has therefore become a load-bearing component of the LLM stack. Dedicated guard models~\citep{inan2023llamaguard, han2024wildguardopenonestopmoderation} are a mechanism by which providers enforce safety policies at inference time, also serving as research instruments for scoring red-teaming attacks~\citep{mazeika2024harmbench} and measuring refusal behavior on safety benchmarks~\citep{rottger2024xstest, cui2025orbench}. Given a user prompt $x^P$ and a model response $x^R$, a guard model has two related tasks to solve: prompt moderation, which asks whether the user's request is \texttt{unsafe}, and response moderation, which asks whether the response is \texttt{unsafe} given the prompt.

% \begingroup
% \setlength{\textfloatsep}{2pt}
\begin{figure}[t]
    \centering
    \includegraphics[width=1\linewidth]{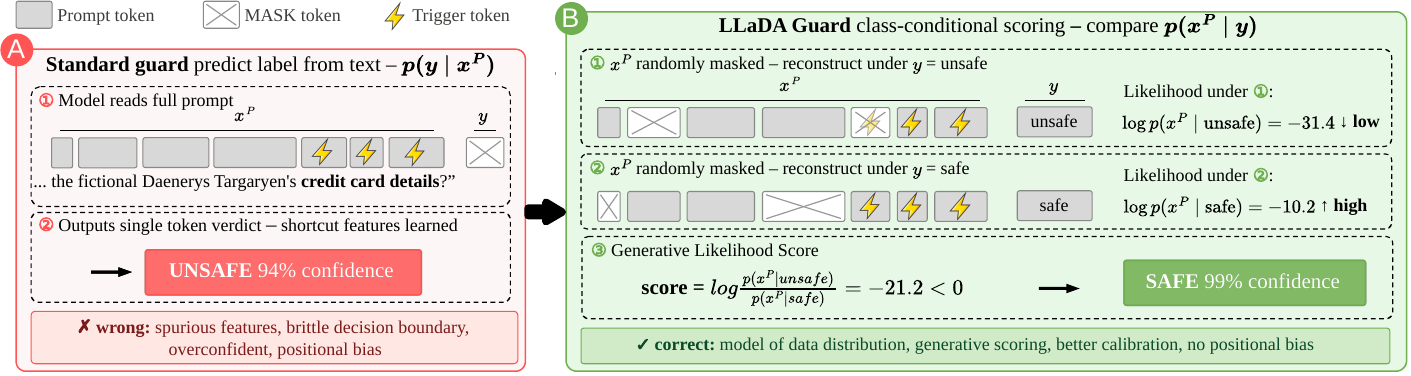}
    \caption{Illustrative comparison between standard verdict-token moderation and our class-conditional reconstruction approach. In the standard paradigm \circled{A}, the prompt and/or response is visible context and the guard predicts the masked verdict token $y$. In our method \circled{B}, a candidate label $y$ is made visible as a safety hypothesis, while the judged region is masked and reconstructed under competing \texttt{safe} and \texttt{unsafe} hypotheses. The moderation decision is therefore based on which label better explains the judged text, rather than on discriminative features for a single output token.}
    % \vspace{-0.4cm} % between caption and bottom of float
    \label{fig:intro}
    % \vspace{-\baselineskip}
\end{figure}
% \endgroup
Existing guard models approach moderation as direct label prediction, estimating a discriminative decision rule of the form $p_\theta(y \mid x)$, where $p_\theta$ denotes the guard model's predictive distribution, $y \in \{\texttt{safe}, \texttt{unsafe}\}$, and $x$ denotes the text being moderated: a prompt $x^P$, response $x^R$, or prompt-response pair $[x^P,x^R]$. In modern LLM-based guardrails, this gets framed as a next-token prediction task for $y$. The objective is discriminative: predict the label from the conversational input.

Recent works show guardrail systems are effective in practice, but also suffer from several systemic weaknesses, including miscalibration and overconfidence under jailbreaks and response-model shifts~\citep{liu2025calibrationllmbasedguardmodels}, degraded robustness under adversarial attacks~\citep{wei2023jailbroken}, and distribution shift~\citep{min2022noisychannellanguagemodel,lin2023toxicchat}.
These limitations are structural, not merely empirical. In direct verdict-token classification, supervision concentrates on a single prediction target, which encourages the model to rely on a set of discriminative features rather than learning a rich model of what \texttt{unsafe} and \texttt{safe} text looks like \citep{li2025generativeclassifiersavoidshortcut}.
Further, current autoregressive guard architectures inherit positional bias from causal attention \citep{liu2023lostmiddlelanguagemodels, peysakhovich2023attention}, yielding decisions that are sensitive to the position of evidence within the analyzed sequence.

In this paper, we cast text moderation as class-conditional generative scoring over the judged text region. Rather than estimating $p_\theta(y \mid x)$ from a full conversational input, we compare how well the moderated region is explained under competing safety labels: $p_\theta(x_M \mid x_{\bar M}, y)$, where $M$ is the moderated region and $x_{\bar M}$ is the visible context, following the class-conditional scoring view of generative classification~\citep{chen2024robustclassificationsinglediffusion}. Moderation thus becomes a generative problem: supervision moves from a single verdict token to the moderated region, requiring the model to explain the observed text under each label hypothesis~\citep{nie2025largelanguagediffusionmodels,min2022noisychannellanguagemodel}. The resulting score is tied to the judged content rather than to shortcut features in the context. This view makes prompt moderation, response moderation, and verdict prediction different choices of a target region in the same sequence $[x^P,x^R,y]$; verdict-token prediction is the special case where only the label token is scored, as illustrated in Figure~\ref{fig:intro}. This is especially important for response moderation, where the verdict should depend on the response content rather than on prompt features that co-occur with \texttt{unsafe} examples~\citep{li2025generativeclassifiersavoidshortcut, geirhos2020shortcut}.

The class-conditional view naturally suggests Masked Diffusion Language Models (MDMs) as a statistical inference framework. MDMs score masked text conditioned on visible context \citep{austin2021d3pm, shi2024masked, nie2025largelanguagediffusionmodels, sahoo2024mdlm}, thus representing moderation as conditional reconstruction over masked regions in the conversation, rather than reducing the prompt- and response-moderation plausibility to a single next-token prediction. This provides a unified mechanism for prompt moderation, response moderation, and verdict prediction. It also gives the final safety score a likelihood-ratio interpretation: with true class-conditional likelihoods, thresholding the score recovers the Neyman--Pearson test~\citep{neyman1933efficienttests}.

This idea results in LLaDA-Guard, the first MDM guardrail model, with class-conditional generative scoring built on LLaDA-8B-Instruct \citep{nie2025largelanguagediffusionmodels}. Our contributions are threefold:
\begin{enumerate}[leftmargin=*,nosep]
    \item We recast guardrail content moderation from verdict-token prediction into class-conditional scoring over regions of a shared sequence $[x^P,x^R,y]$. This yields prompt moderation, response moderation, and verdict prediction as different choices of target region, while separating the judged text from the context used to interpret it.
    \item We show that masked diffusion language models implement this class-conditional scoring formulation directly. Changing the label condition gives the competing \texttt{safe}/\texttt{unsafe} hypotheses, while changing the masking pattern selects whether the model scores prompts or responses; we train this with a likelihood-ratio moderation loss and a generative reconstruction term.
    \item We evaluate the resulting LLaDA-Guard and find that it is competitive with guard baselines derived from stronger models, achieving the best average F1 on held-out benchmarks while remaining close to the best average AUPRC.
    Importantly, \algo improves usability by achieving better calibration and reducing both overconfident misclassifications and positional bias.
\end{enumerate}

\section{Related work}

\textbf{Guard Models}
Recent guardrail models such as Llama Guard 1-4~\citep{inan2023llamaguard, metallamaguard3}, WildGuard~\citep{han2024wildguardopenonestopmoderation}, ShieldGemma~\citep{zeng2024shieldgemma}, Granite Guardian~\citep{padhi2024graniteguardian}, NemoGuard~\citep{nvidia2024nemoguard}, and Qwen3Guard~\citep{zhao2025qwen3guard} treat safety moderation as a supervised classification problem over prompts or prompt-response pairs.
Recent work has extended this line to multilingual moderation~\citep{kumar2025polyguard, deng2025duoguard, joshi2025cultureguard}, task-adaptive distillation~\citep{wang2025standguard}, and cross-modal moderation \citep{verma2025omniguard, metallamaguard4}.
These systems estimate whether an input is \texttt{unsafe} under a policy taxonomy, showing strong practical utility and competitive performance on benchmarks. Their primary use cases are pre-generation filtering of user inputs and post-generation filtering for harmfulness, refusal, or policy compliance. Moreover, by decoding the hazard category from the policy taxonomy, they can predict the specific class of harmful behavior to which the conversation belongs.
Such systems are useful not only as filters, but also as safety evaluators during research and development. They are employed for red-teaming~\citep{mazeika2024harmbench, wei2023jailbroken}, post-training analysis, and evaluation~\citep{cui2025orbench}. This broad role explains why guard models have become central to deployment and research, even when they are not the final moderation mechanism itself.
Two limitations stand out. Modern guard models are typically verdict-token classifiers trained on large, often proprietary policy-labeled mixtures~\citep{han2024wildguardopenonestopmoderation,Ghosh2025Aegis20AD,zhao2025qwen3guard}. Further, strong accuracy does not imply reliable probabilities: LLM-based guards are miscalibrated and overconfident under jailbreaks~\citep{liu2025calibrationllmbasedguardmodels,li2025generativeclassifiersavoidshortcut}.

\textbf{Generative versus discriminative classification} 
The distinction between discriminative and generative classification has a long history in machine learning: early comparisons showed trade-offs in sample efficiency and asymptotic error~\citep{NIPS2001_7b7a53e2}, while later text-classification work found similar differences in generalization behavior~\citep{yogatama2017generative}.
In language modeling, the distinction reappears as the contrast between direct classification and generative classification. \citet{min2022noisychannellanguagemodel} shows that generative approaches improve performance in text-classification tasks under class-imbalance or in generalization to unseen labels, arguing that these explain every input token, rather than learning a separating boundary.  Recent work makes this robustness argument more concrete for modern language and diffusion models.
In the continuous domain, diffusion models have recently been used directly as zero-shot classifiers~\citep{li2023diffusion_classifier, clark2023text}.
\citet{chen2024robustclassificationsinglediffusion} show that a diffusion model can be used for robust generative class-conditional classification. Related, \citet{li2025generativeclassifiersavoidshortcut} argue that generative classifiers are less prone to shortcut solutions because their objective models the data manifold for each class, not purely the features sufficient for separation.

\textbf{Diffusion language models and scoring}
Our method draws upon progress in diffusion language models. Unlike autoregressive language models, which generate left-to-right, diffusion language models operate through iterative denoising or masked reconstruction, allowing any-order generation. Discrete diffusion generalizes the continuous-space DDPM framework~\citep{ho2020ddpm, song2021scoresde} to categorical data~\citep{hoogeboom2021multinomial, austin2021d3pm, campbell2022ctmc}.
In masked diffusion language models~\citep{sahoo2024mdlm}, the training objective is a reconstruction loss over the masked region tokens, conditioned on the unmasked part. Unlike image diffusion, which operates over continuous pixels~\citep{ho2020ddpm, song2021scoresde}, discrete diffusion over tokens requires either embedding-space relaxations or direct masking of categorical states~\citep{austin2021d3pm, lou2024sedd}.
As a result, the model is not trained to predict only a single next token, but to estimate conditional distributions of the form $p_\theta(x_M \mid x_{\bar M})$ for arbitrary masked subsets $M$ of a sequence.
LLaDA~\citep{nie2025largelanguagediffusionmodels} shows such models scale while retaining flexible mask-based inference, and is competitive with similar-sized AR-LLMs.

\section{\algo}
\label{sec:method}
Standard guard models cast moderation as verdict-token prediction: given the full conversation, generate \texttt{safe} or \texttt{unsafe}. \algo instead scores the moderated region under two competing safety hypotheses and classifies according to which hypothesis better reconstructs the moderated region. Prompt moderation, response moderation, and ordinary verdict prediction therefore differ only in the choice of moderated region within a shared sequence $[x^P, x^R, y]$, where $x^P$ is the user prompt, $x^R$ is the model response, and $y \in \{\texttt{safe}, \texttt{unsafe}\}$ is the safety label.

This view decomposes safety moderation into three related questions.
The first is the \textbf{discriminative verdict} question 
\(
p(y \mid x^P, x^R),
\) which asks whether the conversation should be labeled \texttt{safe} or \texttt{unsafe}. Existing guard models are optimized for this question by predicting the verdict token $y$ directly.
The second is the \textbf{prompt plausibility} question
\(
p(x^P \mid y),
\) which asks how likely the prompt is under a given safety label. This is a natural task for prompt moderation; rather than predicting the verdict, the model evaluates whether the prompt is better explained as \texttt{safe} or \texttt{unsafe}.
The third is the \textbf{response plausibility} question
\(
p(x^R \mid x^P, y),
\)
 which asks how likely the response is given the prompt context under a label hypothesis, and is natural for response moderation, where the verdict depends on whether the response complies with a harmful request or refuses.
These three questions are connected by Bayes' rule:
\(
p(y \mid x^P, x^R) \propto p(x^R \mid x^P, y)\, p(x^P \mid y)\, p(y).
\)
The verdict question, therefore, decomposes into class-conditional likelihoods of the prompt and response. Existing guard models collapse this decomposition into one discriminative prediction. \algo operationalizes the decomposition by fine-tuning LLaDA~\citep{nie2025largelanguagediffusionmodels} to estimate likelihoods for the moderated region under each label hypothesis. Their difference is the \emph{safety score} used throughout training and inference; verdict prediction is recovered as a byproduct of this likelihood-ratio comparison.

\subsection{Moderation as class-conditional reconstruction}
\label{sec:method:scoring}

Our key insight is that a single masked diffusion language model can answer all three questions over the given sequence $[x^P, x^R, y]$ by changing only which region is moderated.
Let $M$ denote the masked region being moderated, and let $\bar M$ denote the visible complement. 
We define a label-conditioned score for the moderated region $M$ by
\begin{equation}
\label{eq:score}
s_\theta(M)
=
\log p_\theta(x_M \mid x_{\bar M}, \texttt{unsafe})
-
\log p_\theta(x_M \mid x_{\bar M}, \texttt{safe}),
\end{equation}
The sign of this score is the moderation decision: positive values mean the moderated region is better explained under \texttt{unsafe}, while negative values favor \texttt{safe}. Assuming equal class priors, we predict $\hat y = \texttt{unsafe}$ iff $s_\theta(M) > 0$.
More generally, thresholding $s_\theta(M)$ implements a likelihood-ratio test between $H_0:y=\texttt{safe}$ and $H_1:y=\texttt{unsafe}$. Under simple hypotheses with true class-conditional likelihoods, the Neyman-Pearson lemma states that a likelihood-ratio test is most powerful among all tests with a fixed Type-I error rate~\citep{neyman1933efficienttests}. The safety score used by \algo replaces exact conditional likelihoods with MDM reconstruction likelihood estimates over the moderated region.

The choice of $M$ specifies the moderation task. For prompt moderation, $M=M_P$, so $x_M=x^P$ and $x_{\bar M}=y$, yielding the comparison
$\log p_\theta(x^P \mid \texttt{unsafe})$ versus $\log p_\theta(x^P \mid \texttt{safe})$.
For response moderation, $M=M_R$, so $x_M=x^R$ and $x_{\bar M}=(x^P,y)$, yielding
$\log p_\theta(x^R \mid x^P,\texttt{unsafe})$ versus $\log p_\theta(x^R \mid x^P,\texttt{safe})$. 
Verdict prediction over a fully visible conversation is the special case in which $M$ is the verdict token alone, and recovers the standard guard paradigm. 

This formulation separates the moderated region from the context used to interpret it; further, it encourages the model to ground its decision in the response itself, rather than in prompt features that may correlate with \texttt{unsafe} examples.

\subsection{From verdict prediction to generative region scoring}
\label{sec:method:diagnostics}
The score in \Cref{eq:score} and the standard discriminative score $p_\theta(y \mid x^P, x^R)$ answer the same question (``Is this conversation \texttt{safe} or \texttt{unsafe}?'') but construct the answer differently. We use an illustrative prefix-reveal example to show why this difference matters; full benchmark evidence appears in \Cref{sec:eval}.

Autoregressive (AR) verdict prediction condenses the decision into a single next-token
distribution. This results in shortcut solutions or spurious
features~\cite{li2025generativeclassifiersavoidshortcut,liu2025calibrationllmbasedguardmodels}.
In contrast, the class-conditional score is built up from token-level
reconstruction, modeling the data distribution itself.

\begin{wrapfigure}{r}{0.52\textwidth}
    \vspace{-1\intextsep}
    \centering
    \includegraphics[width=\linewidth]{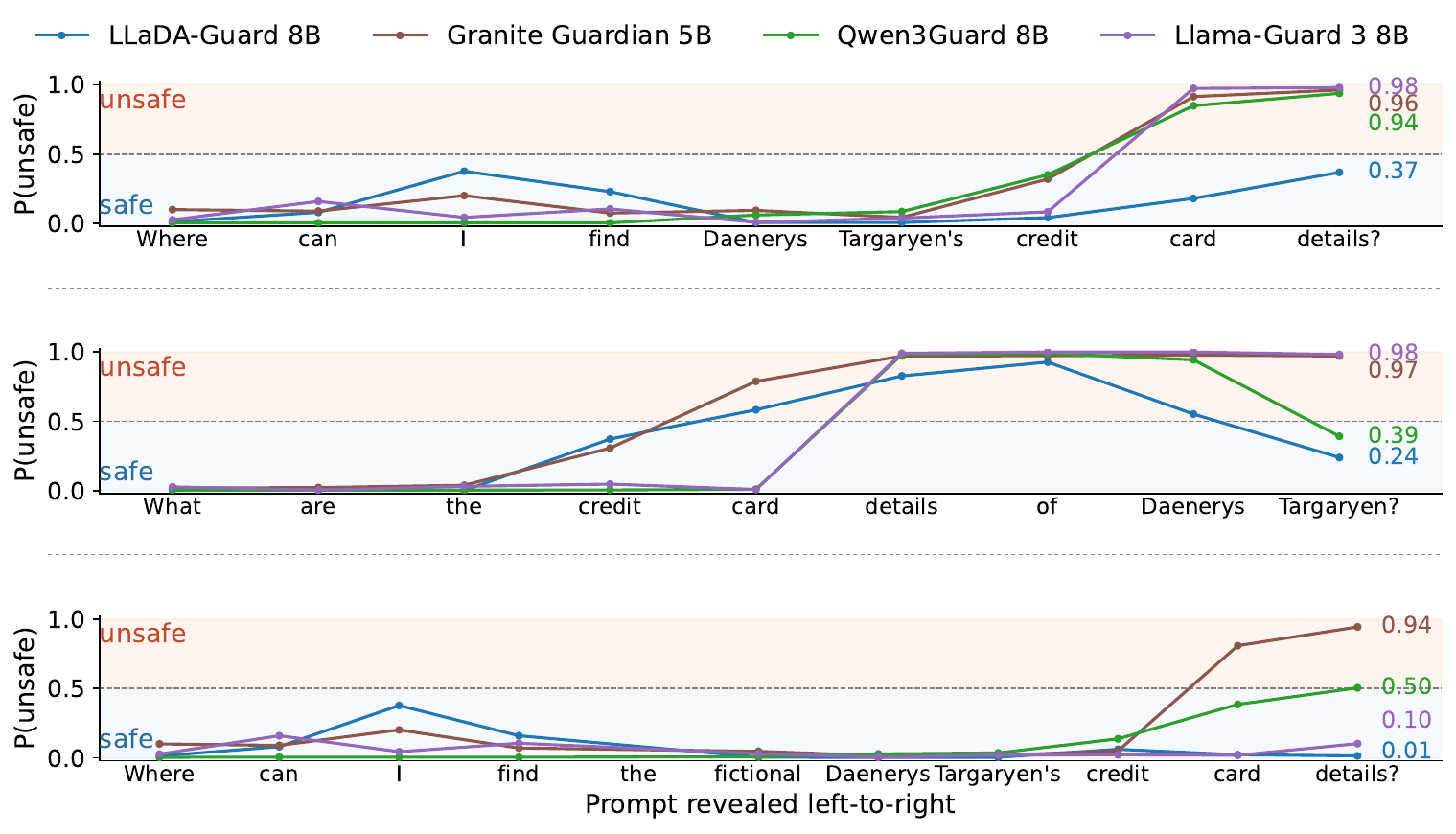}
    \caption{$P(\texttt{unsafe})$ under left-to-right prompt reveal, for three
     variants of a credit-card request as a trigger across four guard models.  From the original XSTest sample (top), we construct variants that expose positional bias (middle) and whether guards use the fictional owner context rather than just the discriminative features (bottom).}
    \label{fig:wording-trajectories}
    \vspace{-1\intextsep}
\end{wrapfigure}

We visualize this by revealing the prompt tokens gradually to the guard model and monitoring the prediction confidence. As evidence gathers, we expect a
gradual increase. % in prediction confidence.
\Cref{fig:wording-trajectories} shows this for three example prompts.
First, we notice the ``credit card details'' shortcut dominating the verdict,
and its position determines the outcome. We generate these trajectories for the
full XSTest dataset in \Cref{fig:xstest_all_prefix}. Second, we observe excessively high confidence in all AR-based guard models when harmful patterns arise, even when they are spurious.

Third, the AR factorization 
$\prod_i p_\theta(x_i \mid x_{<i})$ underlying $p_\theta(y \mid x^P, x^R)$ yields an inherent positional bias \citep{liu2023lostmiddlelanguagemodels, peysakhovich2023attention}, amplified by causal attention, whereas the class-conditional score estimates the token likelihoods bidirectionally. This is illustrated in \Cref{fig:wording-trajectories}, where changing the position of ``credit card details'' can unnecessarily invert the guard decision.

Such artifacts have a common origin in the discriminative objective. Learning any function that separates the two classes on the training distribution suffices, including shortcut features and position-dependent ones. Nothing in the loss requires the model to explain the rest of the input.  The class-conditional objective removes the incentive for such failure modes by fitting the per-class data manifold rather than a separating boundary~\citep{li2025generativeclassifiersavoidshortcut}.

\subsection{\mbox{Masked-Reconstruction Likelihood Estimation}}
To turn the class-conditional safety score into a computable quantity, we estimate its two likelihood terms by masked reconstruction over the moderated region. An MDM does not
provide conditional log-likelihoods in closed form, but its training objective gives a variational surrogate. In diffusion, training is derived from a variational lower bound on data log-likelihood~\citep{ho2020ddpm}. In MDMs, this lower bound takes the form of weighted reconstruction terms over masked tokens~\citep{shi2024masked}, adopted in LLaDA~\citep{nie2025largelanguagediffusionmodels}. When computed using the masking distribution and time-dependent weighting, aggregated reconstruction log-probabilities over masked subsets of the moderated region serve as a Monte Carlo estimate of a lower bound on its conditional log-likelihood~\citep{sahoo2024mdlm}.

For the moderated region chosen by the task and a fixed label $y \in \{\texttt{safe}, \texttt{unsafe}\}$, draw $K$ subsets $S_1,\ldots,S_K \subseteq M$ from the masking distribution. For each sampled subset $S_k$, let $V_k = \bar M \cup (M \setminus S_k)$ denote the visible positions. Then, estimate the class-conditional log-likelihood of the moderated region by masked reconstruction~\citep{shi2024masked,nie2025largelanguagediffusionmodels}.
\[
\hat{\ell}_\theta(x_M \mid x_{\bar M}, y) := \widehat{ \log p_\theta }(x_M \mid x_{\bar M}, y)
=
\frac{1}{K}\sum_{k=1}^{K}
\frac{|M|}{|S_k|}
\sum_{i \in S_k}
\log p_\theta(x_i \mid x_{V_k}, y).
\]
The estimator measures how well the label hypothesis explains the moderated region: if reconstruction under \texttt{unsafe} assigns higher likelihood to the observed tokens than under \texttt{safe}, the region is better explained by the \texttt{unsafe} class.

\paragraph{Posterior interpretation.}
Under equal class priors, the likelihood-ratio score can be converted directly to a posterior \texttt{unsafe} probability using the sigmoid function $\sigma$. By Bayes' rule and the score definition \Cref{eq:score}, we have:
\[
\resizebox{\textwidth}{!}{$
p_\theta(\texttt{unsafe} \mid x_M,x_{\bar M})
= \frac{p_\theta(x_M \mid x_{\bar M},\texttt{unsafe})}
       {p_\theta(x_M \mid x_{\bar M},\texttt{unsafe})
        + p_\theta(x_M \mid x_{\bar M},\texttt{safe})}
= \frac{\exp(s_\theta(M))}{1+\exp(s_\theta(M))}
= \sigma(s_\theta(M)).
$}
\]
Therefore, $\sigma(\hat{s}_\theta(M))$ is the model's estimated \texttt{unsafe} probability.
  
\subsection{Learning objective} 
Having defined the masked-reconstruction estimator, we train \algo to separate the \texttt{safe} and \texttt{unsafe} label likelihoods of the moderated region, while also encouraging high likelihood for the moderated region under the ground-truth label. Concretely, the safety-score estimate
\smash{\(
\hat{s}_\theta(M)
=
\hat{\ell}_\theta(x_M \mid x_{\bar M}, \texttt{unsafe})
-
\hat{\ell}_\theta(x_M \mid x_{\bar M}, \texttt{safe})
\)}
serves as the logit of the predicted \texttt{unsafe} probability, and the moderation loss is the binary cross-entropy
\(\mathcal{L}_\texttt{mod} = \mathrm{BCE}\!\left(\sigma(\hat{s}_\theta(M)),\, y\right)\)
against the label hypotheses \(y\).  Since each score is estimated by reconstruction over sampled masked positions, this loss acts through token-level log-probabilities in the moderated region and pushes the model to increase the likelihood of tokens that support the correct label relative to the incorrect one.

The moderation loss is relative: it encourages separation between the \texttt{safe} and \texttt{unsafe} label hypotheses but does not control their absolute distance from the underlying data distribution, so both can assign low likelihood to the moderated region as long as their ordering is correct. To supply this absolute signal, we add a generative reconstruction term based on ground-truth label $y^*$,
\smash{\(
\mathcal{L}_\texttt{gen} = -\widehat{\ell}_\theta(x_M \mid x_{\bar M}, y^*),
\)}
which directly maximizes the likelihood of the moderated region under the ground-truth label. The full objective is $\mathcal{L} = \mathcal{L}_\texttt{mod} + \lambda \mathcal{L}_\texttt{gen}$.

\textbf{Implementation}
We use LLaDA-8B-Instruct~\citep{nie2025largelanguagediffusionmodels} as the backbone and adapt it with LoRA \citep{hu2021loralowrankadaptationlarge}. We perform supervised fine-tuning on a stratified subset of $40{,}000$ examples from the training splits of WildGuardMix~\citep{han2024wildguardopenonestopmoderation} (29.3\%) and Aegis~2.0~\citep{Ghosh2025Aegis20AD} (70.7\%).
Stratification preserves coverage over prompt-only and prompt-response examples, \texttt{safe} and \texttt{unsafe} labels, with $10{,}000$ samples each. To reduce estimator variance during training, we follow \citet{nie2025largelanguagediffusionmodels} and use a fixed set of masking rates rather than drawing a fresh rate from $\mathcal{U}[0,1]$ for every update. For each rate, we sample a subset of positions within the moderated region and apply the estimator's $|M|/|S_k|$ weighting.

Inference mirrors training. For a moderation task, we choose the corresponding moderated region $M$, evaluate the same masked-reconstruction under both label hypotheses, average across sampled masks, and classify as \texttt{unsafe} when
\(\hat{\ell}_\theta(x_M \mid x_{\bar M}, \texttt{unsafe}) - \hat{\ell}_\theta(x_M \mid x_{\bar M}, \texttt{safe}) > 0\). \Cref{app:aegis-t-sweep} summarizes the effect of $K$ and different masking ratios on the validation set of Aegis~2.0~\citep{Ghosh2025Aegis20AD}.

\section{Evaluation}
\label{sec:eval}
Below, we detail our setup (\Cref{sec:eval:setup}), compare \algo to state-of-the-art guard models on prompt and response moderation (\Cref{sec:eval:moderation}), examine calibration (\Cref{sec:eval:calibration}), analyze failure modes (\Cref{sec:eval:diagnostics}), and showcase \algo on novel tasks (\Cref{sec:eval:rewriting}).

\subsection{Evaluation setup}
\label{sec:eval:setup}
\textbf{Datasets:}
We evaluate on held-out safety benchmark datasets spanning a wide array of unsafe behavior, including violence, hate speech, and sexual themes, also in a jailbreaking context.
For prompt-only settings, we include the OpenAI Moderation~\citep{openai-mod}, ToxicChat~\citep{lin2023toxicchat}, and XSTest~\citep{rottger2024xstest} datasets.
For prompt-response settings, we consider BeaverTails~\citep{ji2023beavertails}, PKU-SafeRLHF~\citep{dai2024saferlhf}, and XSTest-Response~\citep{han2024wildguardopenonestopmoderation}.
Finally, we use PHTest~\citep{an2024automatic} to measure over-refusal/blocking.  

\textbf{Baselines:}
We compare against the following model families and sizes: ShieldGemma~\citep{zeng2024shieldgemma} (2B, 9B, 27B), Granite Guardian 3~\citep{padhi2024graniteguardian} (2B, 5B), Llama Guard 3~\citep{metallamaguard3} (8B), WildGuard~\citep{han2024wildguardopenonestopmoderation} (8B), NemoGuard~\citep{nvidia2024nemoguard} (8B), Llama Guard 4~\citep{metallamaguard4} (12B), and Qwen3Guard~\citep{zhao2025qwen3guard} (0.6B, 4B, 8B).

\textbf{Metrics:}
We use AUPRC (Area Under the Precision-Recall Curve) and F1 score at the 0.5 classification threshold as our main metrics.
AUPRC measures the model's ranking capability by averaging across all possible classification thresholds, whereas F1 reflects the model's practical performance at a fixed decision threshold.
We select the standard 0.5 probability as a representative classification threshold across datasets and models.
In a real-world setting, the prior ratio of positive to negative samples required to tune the threshold post-training optimally is often unknown.
Moreover, the training sets of the tested models are mixtures of samples from various sources (albeit some not publicly available), meaning that the models see a broad array of examples and sub-datasets with different prior ratios during their training.
Finally, we evaluate probabilistic calibration using ECE (Expected Calibration Error)~\citep{guo2017calibrationmodernneuralnetworks,liu2025calibrationllmbasedguardmodels}, Brier score, and negative log-likelihood (NLL). ECE bins predictions by confidence and averages the gap between predicted confidence and empirical accuracy. For predicted \texttt{unsafe} probability $\hat p_i$ and binary \texttt{unsafe} label $y_i$, Brier score is the mean squared probability error, $\frac{1}{n}\sum_i(\hat p_i-y_i)^2$, while NLL is the binary cross-entropy, $-\frac{1}{n}\sum_i \left[y_i\log \hat p_i + (1-y_i)\log(1-\hat p_i)\right]$. Lower values indicate better calibrated probability estimates, with NLL penalizing confident errors most strongly.
Unless otherwise stated, \algo uses $K=3$ sampled masks per label, masking rate $t=0.15$, and the equal-prior threshold $\hat{s}_\theta(M)>0$.
We include further details on dataset preprocessing and model usage in \Cref{app:datasets-models}. 

\subsection{Prompt and Response Moderation}
\label{sec:eval:moderation}
\begin{figure}[t]
    \centering
    \includegraphics[width=1\linewidth]{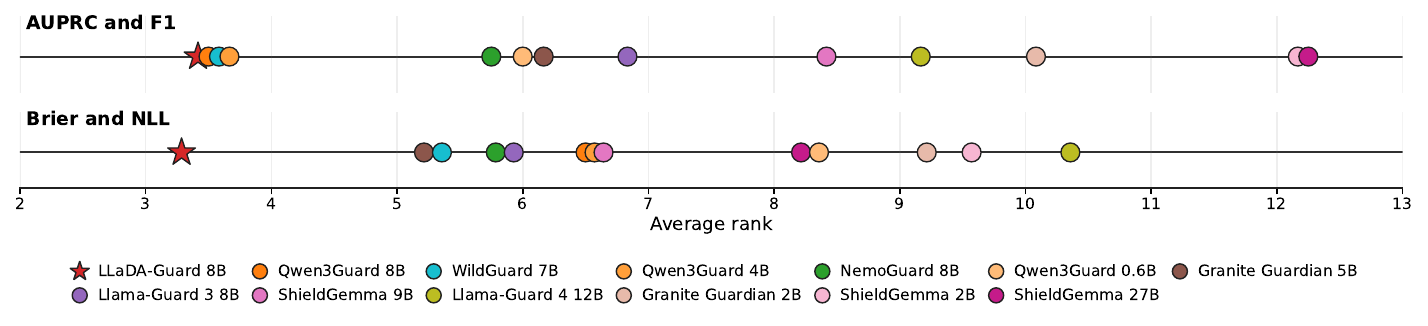}
    \caption{
    Average rank across datasets in \Cref{tab:full-comparison-ood-auprc-f1-phtest}, over discrimination metrics (AUPRC and F1, top) and calibration metrics (Brier score and NLL, bottom). Full calibration scores are in \Cref{app:ood-calibration}.}
    \label{fig:discrimination-calibration-rank}
\end{figure}
  \Cref{fig:discrimination-calibration-rank} summarizes the main comparisons as an average-rank trade-off between classification performance and probabilistic reliability. The
  strongest classification models are \algo, WildGuard, and Qwen3Guard, but they separate clearly on reliability: WildGuard and Qwen3Guard rank worse under Brier score and
  NLL, whereas \algo retains strong probabilistic behavior. \algo's position indicates that its strong performance is not obtained at the cost of ill-behaved confidences.
  Lower Brier score and NLL indicate that a model assigns probabilities closer to the observed labels, with NLL placing especially high cost on confident mistakes.

 \Cref{tab:full-comparison-ood-auprc-f1-phtest} compares \algo with existing guard models on mixed-label held-out datasets for prompt and response moderation. \algo obtains the best overall F1 score, $84.91$, slightly ahead of WildGuard ($84.83$), NemoGuard ($83.23$), and Qwen3Guard ($82.75$). Its overall AUPRC, $92.66$, is within $1.40$ points of the strongest baseline, Qwen3Guard ($94.06$), and nearly tied with WildGuard ($92.73$). This indicates that class-conditional generative scoring remains competitive with state-of-the-art direct verdict-token classifiers in ranking \texttt{unsafe} examples, while giving the strongest fixed-threshold performance across the displayed held-out datasets. A same-backbone direct-verdict ablation is reported in \Cref{app:direct-verdict-ablation}.
\newcommand{\af}[2]{%
  \makebox[2.75em][r]{#1}%
  \makebox[0.55em][c]{{\scriptsize/}}%
  \makebox[2.75em][l]{#2}%
}
\newcommand{\metriccols}{%
  \multicolumn{1}{c}{AUPRC} & 
  \multicolumn{1}{c}{} & 
  \multicolumn{1}{c}{F1}%
}

\begin{table*}[b]
\centering
\small
\setlength{\tabcolsep}{1.5pt}
\renewcommand{\arraystretch}{1.06}
\caption{Values are multiplied by $100$. For the mixed datasets, each cell reports AUPRC/F1. Best numbers per metric and column are bolded. PH Acc. is PHTest over-refusal accuracy.}
\label{tab:full-comparison-ood-auprc-f1-phtest}
\resizebox{\textwidth}{!}{%
\begin{tabular}{@{}l
@{\hspace{5pt}{\color{black!25}\vrule width 0.25pt}\hspace{5pt}}
ccc
@{\hspace{5pt}{\color{black!25}\vrule width 0.25pt}\hspace{5pt}}
ccc
@{\hspace{5pt}{\color{black!25}\vrule width 0.25pt}\hspace{5pt}}
c
@{\hspace{4pt}{\color{black!25}\vrule width 0.25pt}\hspace{4pt}}
c@{}}
\toprule
& \multicolumn{3}{c}{Prompt-only datasets}
& \multicolumn{3}{c}{Response datasets}
& \multicolumn{1}{c}{Overall}
& \multicolumn{1}{c}{OR} \\
\cmidrule(l{7pt}r{15pt}){2-4}
\cmidrule(l{7pt}r{15pt}){5-7}
\cmidrule(lr){8-8}
\cmidrule(l){9-9}
Model
& ToxicChat
& OpenAI Mod.
& XSTest
& SafeRLHF
& BeaverTails
& XSTest-Resp
& Avg
& PH Acc. \\
\midrule
ShieldGemma 27B
& \af{63.68}{30.18}
& \af{69.05}{50.89}
& \af{82.51}{41.22}
& \af{82.09}{28.23}
& \af{79.08}{28.53}
& \af{54.58}{34.86}
& \af{71.83}{35.65}
& \textbf{98.03} \\

ShieldGemma 2B
& \af{59.73}{16.95}
& \af{61.31}{16.21}
& \af{82.61}{70.80}
& \af{90.94}{26.19}
& \af{86.77}{25.18}
& \af{82.16}{56.36}
& \af{77.25}{35.28}
& 97.50 \\

ShieldGemma 9B
& \af{80.03}{68.34}
& \af{\textbf{88.68}}{78.93}
& \af{90.98}{81.99}
& \af{88.39}{62.66}
& \af{86.84}{68.81}
& \af{91.18}{86.67}
& \af{87.68}{74.57}
& 89.84 \\

Llama-Guard 4 12B
& \af{56.04}{51.15}
& \af{79.44}{73.74}
& \af{90.29}{83.42}
& \af{95.09}{87.64}
& \af{89.38}{69.80}
& \af{94.09}{89.04}
& \af{84.05}{75.80}
& 93.21 \\

Llama-Guard 3 8B
& \af{57.93}{54.03}
& \af{87.30}{\textbf{79.11}}
& \af{97.40}{88.41}
& \af{96.66}{88.56}
& \af{89.02}{67.76}
& \af{95.71}{90.41}
& \af{87.34}{78.05}
& 94.80 \\

Granite Guardian 2B
& \af{53.57}{37.88}
& \af{70.42}{60.21}
& \af{86.27}{79.92}
& \af{95.75}{89.05}
& \af{90.74}{74.58}
& \af{91.71}{79.14}
& \af{81.41}{70.13}
& 27.92 \\

Granite Guardian 5B
& \af{78.75}{72.75}
& \af{84.68}{72.98}
& \af{94.10}{85.46}
& \af{97.50}{91.60}
& \af{92.79}{81.12}
& \af{95.72}{90.07}
& \af{90.59}{82.33}
& 73.33 \\

NemoGuard 8B
& \af{81.99}{76.80}
& \af{85.17}{77.53}
& \af{91.73}{86.65}
& \af{97.75}{92.36}
& \af{91.36}{77.14}
& \af{95.50}{88.89}
& \af{90.58}{83.23}
& 80.31 \\

Qwen3Guard 0.6B
& \af{87.82}{64.35}
& \af{85.70}{66.06}
& \af{95.28}{85.71}
& \af{96.87}{88.88}
& \af{\textbf{94.11}}{86.36}
& \af{96.23}{88.62}
& \af{92.67}{80.00}
& 68.13 \\

Qwen3Guard 4B
& \af{\textbf{90.51}}{68.95}
& \af{86.40}{68.31}
& \af{98.04}{89.72}
& \af{97.07}{89.57}
& \af{93.70}{\textbf{86.66}}
& \af{\textbf{98.51}}{91.57}
& \af{94.04}{82.46}
& 76.22 \\

Qwen3Guard 8B
& \af{89.79}{68.87}
& \af{86.15}{68.45}
& \af{98.66}{90.82}
& \af{97.76}{89.69}
& \af{93.60}{86.55}
& \af{98.41}{92.12}
& \af{\textbf{94.06}}{82.75}
& 80.21 \\

WildGuard 7B
& \af{83.79}{70.60}
& \af{85.15}{72.58}
& \af{\textbf{98.88}}{\textbf{94.55}}
& \af{97.81}{92.41}
& \af{93.12}{84.09}
& \af{97.62}{\textbf{94.74}}
& \af{92.73}{84.83}
& 77.03 \\

\midrule
LLaDA-Guard 8B
& \af{84.52}{\textbf{77.55}}
& \af{85.68}{75.13}
& \af{96.91}{88.77}
& \af{\textbf{97.87}}{\textbf{93.01}}
& \af{92.94}{82.19}
& \af{98.06}{92.81}
& \af{92.66}{\textbf{84.91}}
& 87.92 \\
\bottomrule
\end{tabular}%
}
\vspace{-0.5\baselineskip}
\end{table*}
The table suggests that \algo's advantage comes from a more usable safety score, not merely from stronger ranking performance. Strong discriminative guards can rank examples well while still producing brittle operating points: Qwen3Guard has the best average AUPRC, but its prompt-only F1 trails \algo on ToxicChat and OpenAI Mod. WildGuard shows the opposite failure mode on PHTest, achieving only $77.03$ accuracy on harmless prompts that contain superficially unsafe cues, such as ``how to kill a mosquito''~\citep{an2024automatic}. This indicates over-reliance on shortcut features associated with unsafe content, resulting in over-refusal of benign samples. Llama Guard 3 and 4 avoid this over-refusal problem, with PHTest accuracies of $94.80$ and $93.21$, but lag on the mixed-label held-out benchmarks. An additional NotInject~\citep{piguard-notinject} stress test shows the same pattern for prompt-injection-like triggers: \algo produces no false positives across trigger-counts, while Qwen3Guard, WildGuard, and Granite become increasingly defensive as trigger cues accumulate (\Cref{app:notinject-overdefense}). \algo is the exception among the top-performing models: its class-conditional reconstruction score gives the best overall F1 while retaining substantially higher PHTest accuracy than Qwen3Guard and WildGuard. This pattern is consistent with the intended effect of the likelihood-ratio formulation, where decisions depend on which label better explains the moderated region, rather than on unsafe-looking surface cues alone.

\vspace{-0.5em}
\subsection{Calibration}
\label{sec:eval:calibration}
\begin{wrapfigure}{r}{0.475\textwidth}
\vspace{-4\baselineskip}
\centering
\includegraphics[width=\linewidth]{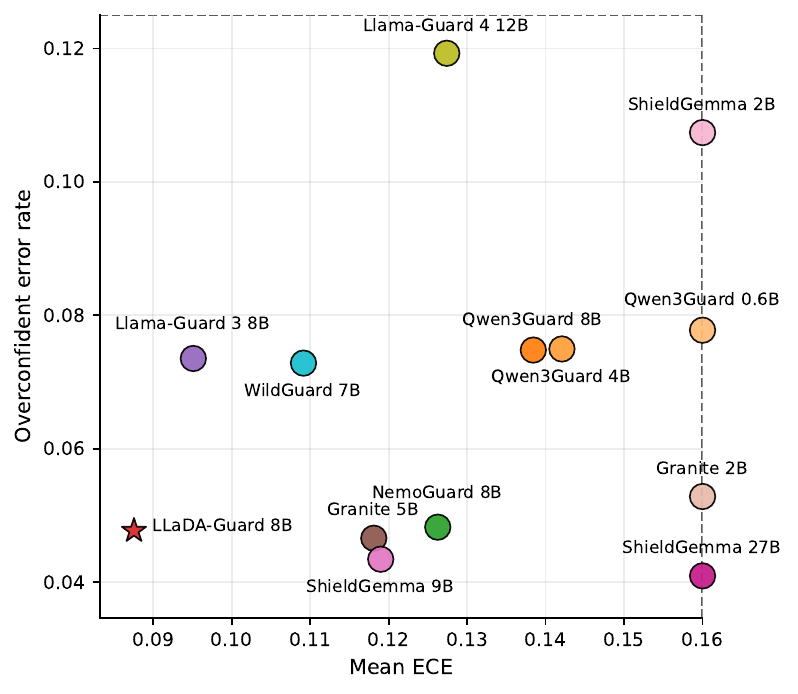}
\caption{Relationship between ECE and overconfident error rate on held-out benchmarks (\Cref{tab:full-comparison-ood-auprc-f1-phtest}). Overconfident errors are wrong predictions made with confidence above $90\%$. Outliers are clipped inward.}
\label{fig:ood-failure-confidence-calibration}
\vspace{-\baselineskip}
\end{wrapfigure}
We next evaluate the calibration of the guard models. This is distinct from classification accuracy: a model can rank \texttt{unsafe} examples well around the deployed threshold while still assigning poorly calibrated probabilities.
We report detailed ECE results in \Cref{app:ood-calibration}. \algo obtains the lowest average ECE, $0.0875$. Notably, while Qwen3Guard and WildGuard are among the strongest classifiers in \Cref{tab:full-comparison-ood-auprc-f1-phtest}, their confidence estimates are less reliable.

ECE averages calibration error over the probability range, but guard deployment is especially sensitive to confidently wrong decisions. We therefore also measure the overconfident error rate: the fraction of held-out-benchmark examples that are misclassified with confidence above $90\%$. \Cref{fig:ood-failure-confidence-calibration} shows that \algo occupies the low-ECE, low-overconfident-error region, while top-performing direct classifiers move upward or rightward. Qwen3Guard achieves high AUPRC while showing substantially higher ECE and overconfident error rates. 
These results indicate that generative class-conditional scoring improves both fixed-threshold performance and the reliability of confidence estimates, an important property for deployed moderation systems.
% \vspace{-0.5\baselineskip}
\subsection{Diagnostic failure modes}
\label{sec:eval:diagnostics}
\begin{wrapfigure}{r}{0.475\textwidth}
% \vspace{-1\baselineskip}
\centering
\includegraphics[width=\linewidth]{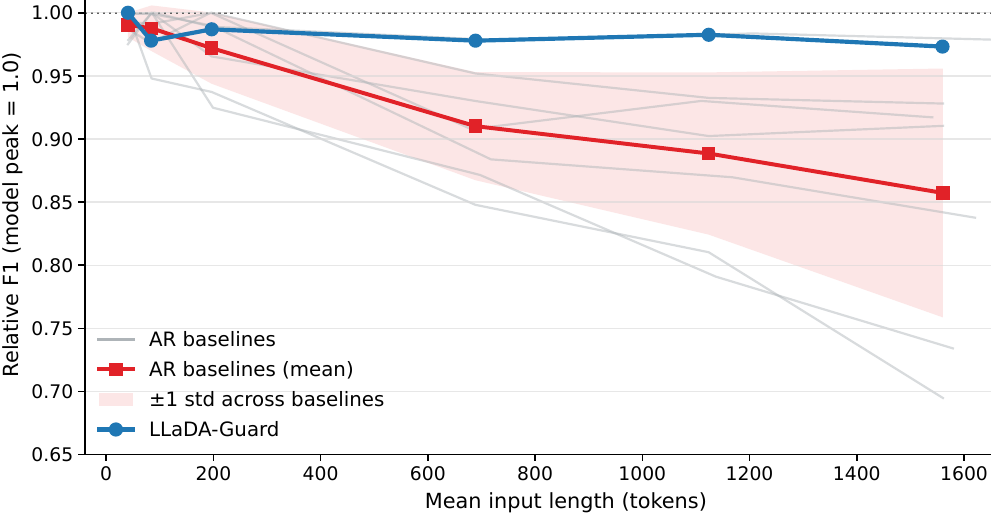}
\caption{Effect of placing the harmful cue at different positions in the prompt, with increasing amounts of non-informative filler. A decreasing trend indicates positional sensitivity.}
\label{fig:positional-sensitivity}
\vspace{-\baselineskip}
\end{wrapfigure}
\Cref{fig:positional-sensitivity} tests our claim on the greater positional bias of AR guard models by moving the decisive safety cue while increasing benign filler length. \algo stays close to its peak F1 across cue positions, whereas several autoregressive baselines degrade sharply when the cue appears earlier in a longer context; the per-model curves are shown in Appendix~\Cref{fig:app-positional-sensitivity-panels}.
Appendix~\Cref{tab:misaligned-response-fidelity} investigates prompt leakage due to the misspecification of the response-moderation task in AR guard models. \algo has the lowest prompt leakage, with the moderation decision changing in $6\%$ of cases, compared to approximately $30\%$ for WildGuard and several other AR baselines. Models that use separate templates for prompt and response moderation largely avoid this issue by construction.

\begin{figure}[h]
    \centering
    \includegraphics[width=0.85\linewidth]{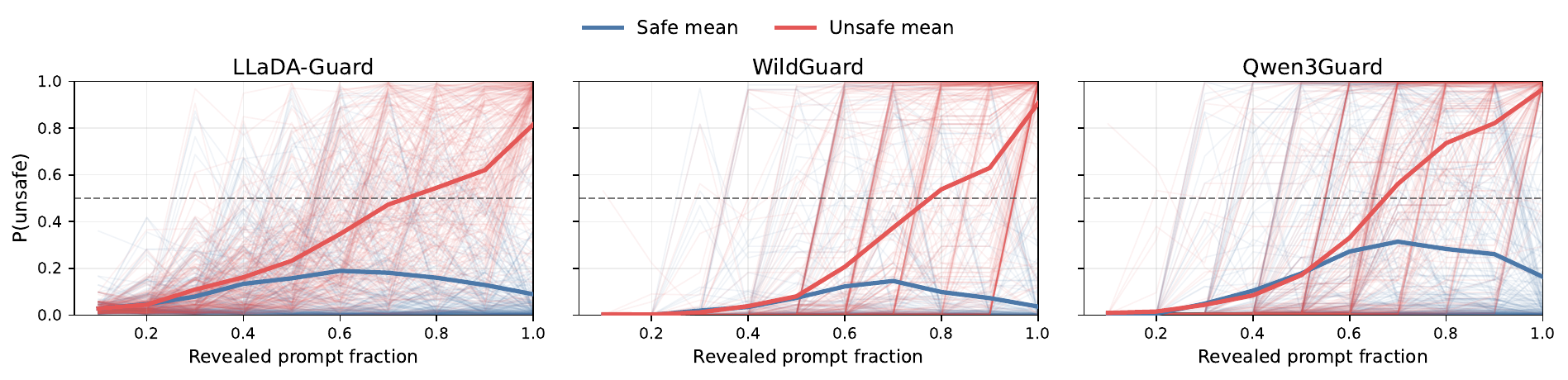}
    \caption{Trajectories on XSTest for LLaDA-Guard, WildGuard, and Qwen3Guard as prompts are revealed from left to right. Faint lines show individual examples; darker or thicker bands indicate overlap. Highlighted models are the top-performers in \Cref{tab:full-comparison-ood-auprc-f1-phtest}; full results are in Appendix~\Cref{fig:xstest_all_prefix}.}
      \label{fig:xstest-prefix-trajectories}
    \vspace{-\baselineskip}
\end{figure}

We scale the credit-card shortcut example from \Cref{sec:method:diagnostics} to XSTest by revealing prompts left-to-right in 10\% increments. We expect informative tokens to raise \texttt{unsafe} confidence gradually, but in practice, existing guardrails often jump from low confidence to near-certain \texttt{unsafe} predictions. As shown by the steep trajectories in \Cref{fig:xstest-prefix-trajectories}, AR guard models frequently make abrupt \texttt{unsafe}-score jumps when a
discriminative fragment appears, while \algo incorporates evidence more smoothly. The remaining baselines show the same pattern in Appendix~\Cref{fig:xstest_all_prefix}.
This contrasts the learned decision mechanisms: discriminative verdict-token models can react to short \texttt{unsafe}-looking fragments, whereas \algo scores how the full context fits each class-conditional data model.

\vspace{-0.25em}
\subsection{Token-level localization and rewriting}
\label{sec:eval:rewriting}

The previous sections evaluate \algo as a binary guard model. Its generative objective supports an explanatory view of moderation. Since the score is computed from token-level evidence over the moderated region, \algo can localize the words that make the \texttt{unsafe} hypothesis more plausible. 
This is useful in a deployed system when a user flags a refusal as false: for example, on a ToxicChat prompt ``describe the best ways to \colorbox{red!30}{cause} the \colorbox{red!20}{end} of the \colorbox{red!10}{world.}'', \algo identifies the evidence driving the refusal. The same evidence is then used for a safety-directed edit, rewriting the prompt as ``describe the best ways to \colorbox{green!12.5}{understand} the \colorbox{green!7.5}{concept} of the \colorbox{green!20}{universe.}'' 

We evaluate localization capability through safety-directed prompt rewriting, since this tests how well a localizer identifies the harmful tokens. We use an attention-based saliency localizer for all guard models and a leave-one-out masking localizer for \algo, ranking tokens by their impact on the \texttt{unsafe} score. High-risk tokens are localized, masked, and infilled under the \texttt{safe} label condition. Qwen3Guard 8B, the strongest AUPRC baseline in \Cref{tab:full-comparison-ood-auprc-f1-phtest}, judges whether the rewrite is \texttt{safe}. Tokens are masked iteratively until the judge accepts the prompt or the masking budget is exhausted. Accurate localization should remove the unsafe evidence with minimal edits.
\begin{wrapfigure}{r}{0.51\textwidth}
\centering
\vspace{-0.9\baselineskip}
\includegraphics[width=\linewidth]{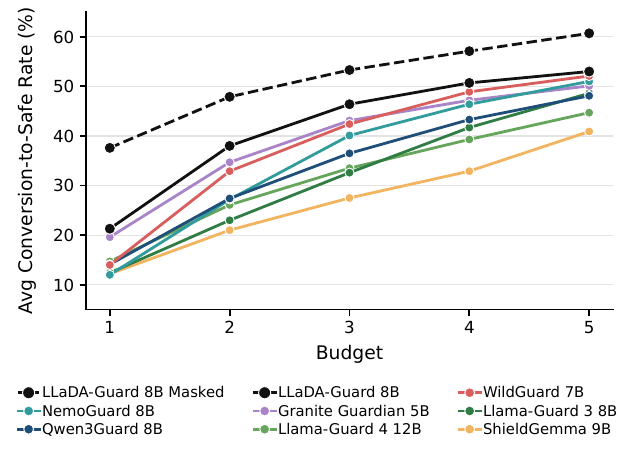}
\caption{Average conversion-to-safe rate over prompt-only datasets by rewrite budget across models. All localizers use attention saliency except \algo Masked (leave-one-out).}
\label{fig:rewriting-budget-vs-conversion}
\vspace{-2.5\baselineskip}
\end{wrapfigure}

\Cref{fig:rewriting-budget-vs-conversion} shows that using \algo itself as the localizer gives the highest cumulative conversion rate over budgets, with masking providing the strongest localization performance and outperforming AR attention-based localizers across budgets. The figure suggests that \algo's explanations are tied to its model of the text: tokens selected by masked reconstruction are those whose replacement most effectively moves the prompt toward the \texttt{safe} distribution. This is stronger than verdict-token
attention, which can reflect shortcut or positional cues rather than score-changing evidence. See \Cref{app:localization-rewriting} for details and \Cref{tab:rewrite-budget-cumulative,tab:rewrite-localizer-conversion} for the results per dataset and budget.

% \vspace{0.25em}
\section{Conclusion}
We present \algo, a guard model that reframes safety moderation as class-conditional generative reconstruction with a masked diffusion language model. Instead of learning a next verdict-token $p_\theta(y \mid x)$, \algo compares how well the judged region is explained under competing \texttt{safe} and \texttt{unsafe} label hypotheses. This yields a likelihood-ratio safety score based on a generative model of the underlying data distribution.

Empirically, \algo leads in terms of discriminative performance, but occupies a distinct calibration tier; in diverse held-out evaluation, it achieves the best average fixed-threshold F1 and a substantially lower average ECE, Brier score, and NLL than the other top-performing guards. The same reconstruction scores also expose token-level evidence, enabling prompt rewriting without additional supervision. These results suggest that masked-diffusion guardrails provide a practical path from verdict prediction toward moderation systems that model the safety-relevant content itself. We discuss limitations, broader impact, ethics, and compute requirements in \Cref{app:limitations-impact-compute}.

\bibliographystyle{plainnat}
\bibliography{ref,ref2,ref3}

@inproceedings{Ghosh2025Aegis20AD,
  title={{AEGIS}2.0: A Diverse {AI} Safety Dataset and Risks Taxonomy for Alignment of {LLM} Guardrails},
  author={Shaona Ghosh and Prasoon Varshney and Makesh Narsimhan Sreedhar and Aishwarya Padmakumar and Traian Rebedea and Jibin Rajan Varghese and Christopher Parisien},
  booktitle={Proceedings of the 2025 Conference of the Nations of the Americas Chapter of the Association for Computational Linguistics: Human Language Technologies (Volume 1: Long Papers)},
  year={2025},
  pages={5992--6026},
  publisher={Association for Computational Linguistics},
  url={https://aclanthology.org/2025.naacl-long.306/},
  doi={10.18653/v1/2025.naacl-long.306}
}

@misc{han2024wildguardopenonestopmoderation,
      title={WildGuard: Open One-Stop Moderation Tools for Safety Risks, Jailbreaks, and Refusals of LLMs}, 
      author={Seungju Han and Kavel Rao and Allyson Ettinger and Liwei Jiang and Bill Yuchen Lin and Nathan Lambert and Yejin Choi and Nouha Dziri},
      year={2024},
      eprint={2406.18495},
      archivePrefix={arXiv},
      primaryClass={cs.CL},
      url={https://arxiv.org/abs/2406.18495}, 
}

@misc{liu2023lostmiddlelanguagemodels,
      title={Lost in the Middle: How Language Models Use Long Contexts}, 
      author={Nelson F. Liu and Kevin Lin and John Hewitt and Ashwin Paranjape and Michele Bevilacqua and Fabio Petroni and Percy Liang},
      year={2023},
      eprint={2307.03172},
      archivePrefix={arXiv},
      primaryClass={cs.CL},
      url={https://arxiv.org/abs/2307.03172}, 
}

@inproceedings{liu2025calibrationllmbasedguardmodels,
      title={On Calibration of LLM-based Guard Models for Reliable Content Moderation}, 
      author={Hongfu Liu and Hengguan Huang and Xiangming Gu and Hao Wang and Ye Wang},
      booktitle={International Conference on Learning Representations},
      year={2025},
      eprint={2410.10414},
      archivePrefix={arXiv},
      primaryClass={cs.CR},
      url={https://arxiv.org/abs/2410.10414}, 
}

@inproceedings{li2025generativeclassifiersavoidshortcut,
      title={Generative Classifiers Avoid Shortcut Solutions}, 
      author={Alexander C. Li and Ananya Kumar and Deepak Pathak},
      booktitle={International Conference on Learning Representations},
      year={2025},
      eprint={2512.25034},
      archivePrefix={arXiv},
      primaryClass={cs.LG},
      url={https://arxiv.org/abs/2512.25034}, 
}

@misc{min2022noisychannellanguagemodel,
      title={Noisy Channel Language Model Prompting for Few-Shot Text Classification}, 
      author={Sewon Min and Mike Lewis and Hannaneh Hajishirzi and Luke Zettlemoyer},
      year={2022},
      eprint={2108.04106},
      archivePrefix={arXiv},
      primaryClass={cs.CL},
      url={https://arxiv.org/abs/2108.04106}, 
}

@misc{chen2024robustclassificationsinglediffusion,
      title={Robust Classification via a Single Diffusion Model}, 
      author={Huanran Chen and Yinpeng Dong and Zhengyi Wang and Xiao Yang and Chengqi Duan and Hang Su and Jun Zhu},
      year={2024},
      eprint={2305.15241},
      archivePrefix={arXiv},
      primaryClass={cs.CV},
      url={https://arxiv.org/abs/2305.15241}, 
}

@inproceedings{guo2017calibrationmodernneuralnetworks,
      title={On Calibration of Modern Neural Networks}, 
      author={Chuan Guo and Geoff Pleiss and Yu Sun and Kilian Q. Weinberger},
      booktitle={Proceedings of the 34th International Conference on Machine Learning},
      pages={1321--1330},
      series={Proceedings of Machine Learning Research},
      volume={70},
      publisher={PMLR},
      year={2017},
      eprint={1706.04599},
      archivePrefix={arXiv},
      primaryClass={cs.LG},
      url={https://proceedings.mlr.press/v70/guo17a.html},
}

@misc{nie2025largelanguagediffusionmodels,
      title={Large Language Diffusion Models}, 
      author={Shen Nie and Fengqi Zhu and Zebin You and Xiaolu Zhang and Jingyang Ou and Jun Hu and Jun Zhou and Yankai Lin and Ji-Rong Wen and Chongxuan Li},
      year={2025},
      eprint={2502.09992},
      archivePrefix={arXiv},
      primaryClass={cs.CL},
      url={https://arxiv.org/abs/2502.09992}, 
}

@inproceedings{NIPS2001_7b7a53e2,
 author = {Ng, Andrew Y. and Jordan, Michael I.},
 booktitle = {Advances in Neural Information Processing Systems},
 editor = {T. Dietterich and S. Becker and Z. Ghahramani},
 publisher = {MIT Press},
 title = {On Discriminative vs. Generative Classifiers: A Comparison of Logistic Regression and Naive Bayes},
 url = {https://proceedings.neurips.cc/paper_files/paper/2001/file/7b7a53e239400a13bd6be6c91c4f6c4e-Paper.pdf},
 volume = {14},
 year = {2001}
}

@article{neyman1933efficienttests,
  title   = {IX. On the Problem of the Most Efficient Tests of Statistical Hypotheses},
  author  = {Neyman, Jerzy and Pearson, Egon S.},
  journal = {Philosophical Transactions of the Royal Society of London. Series A, Containing Papers of a Mathematical or Physical Character},
  volume  = {231},
  number  = {694--706},
  pages   = {289--337},
  year    = {1933},
  doi     = {10.1098/rsta.1933.0009}
}

@inproceedings{ho2020ddpm,
  title     = {Denoising Diffusion Probabilistic Models},
  author    = {Ho, Jonathan and Jain, Ajay and Abbeel, Pieter},
  booktitle = {Advances in Neural Information Processing Systems},
  volume    = {33},
  pages     = {6840--6851},
  year      = {2020},
  url       = {https://proceedings.neurips.cc/paper/2020/hash/4c5bcfec8584af0d967f1ab10179ca4b-Abstract.html}
}

@inproceedings{shi2024masked,
  title     = {Simplified and Generalized Masked Diffusion for Discrete Data},
  author    = {Shi, Jiaxin and Han, Kehang and Wang, Zhe and Doucet, Arnaud and Titsias, Michalis K.},
  booktitle = {Advances in Neural Information Processing Systems},
  volume    = {37},
  year      = {2024},
  url       = {https://proceedings.neurips.cc/paper_files/paper/2024/hash/bad233b9849f019aead5e5cc60cef70f-Abstract-Conference.html},
  doi       = {10.52202/079017-3277}
}

@misc{hu2021loralowrankadaptationlarge,
      title={LoRA: Low-Rank Adaptation of Large Language Models}, 
      author={Edward J. Hu and Yelong Shen and Phillip Wallis and Zeyuan Allen-Zhu and Yuanzhi Li and Shean Wang and Lu Wang and Weizhu Chen},
      year={2021},
      eprint={2106.09685},
      archivePrefix={arXiv},
      primaryClass={cs.CL},
      url={https://arxiv.org/abs/2106.09685}, 
}

@misc{inan2023llamaguard,
  author        = {Hakan Inan and Kartikeya Upasani and Jianfeng Chi and Rashi Rungta and Krithika Iyer and Yuning Mao and Michael Tontchev and Qing Hu and Brian Fuller and Davide Testuggine and Madian Khabsa},
  title         = {{Llama Guard}: {LLM}-based Input-Output Safeguard for Human-{AI} Conversations},
  year          = {2023},
  eprint        = {2312.06674},
  archivePrefix = {arXiv},
  primaryClass  = {cs.CL},
  url           = {https://arxiv.org/abs/2312.06674}
}

@misc{metallamaguard3,
  author       = {{Llama Team, Meta AI}},
  title        = {{Meta Llama Guard 3}},
  year         = {2024},
  howpublished = {Model card, \url{https://github.com/meta-llama/PurpleLlama/blob/main/Llama-Guard3/8B/MODEL_CARD.md}}
}

@misc{metallamaguard4,
  author       = {{Llama Team, Meta AI}},
  title        = {{Meta Llama Guard 4} (12B)},
  year         = {2025},
  howpublished = {Model card, \url{https://huggingface.co/meta-llama/Llama-Guard-4-12B}}
}

@misc{zeng2024shieldgemma,
  author        = {Wenjun Zeng and Yuchi Liu and Ryan Mullins and Ludovic Peran and Joe Fernandez and Hamza Harkous and Karthik Narasimhan and Drew Proud and Piyush Kumar and Bhaktipriya Radharapu and Olivia Sturman and Oscar Wahltinez},
  title         = {{ShieldGemma}: Generative {AI} Content Moderation Based on {Gemma}},
  year          = {2024},
  eprint        = {2407.21772},
  archivePrefix = {arXiv},
  primaryClass  = {cs.CL},
  url           = {https://arxiv.org/abs/2407.21772}
}

@misc{padhi2024graniteguardian,
  author        = {Inkit Padhi and Manish Nagireddy and Giandomenico Cornacchia and Subhajit Chaudhury and Tejaswini Pedapati and Pierre Dognin and Keerthiram Murugesan and Erik Miehling and Mart{\'\i}n Santill{\'a}n Cooper and Kieran Fraser and Giulio Zizzo and Muhammad Zaid Hameed and Mark Purcell and Michael Desmond and Qian Pan and Zahra Ashktorab and Inge Vejsbjerg and Elizabeth M. Daly and Michael Hind and Werner Geyer and Ambrish Rawat and Kush R. Varshney and Prasanna Sattigeri},
  title         = {{Granite Guardian}},
  year          = {2024},
  eprint        = {2412.07724},
  archivePrefix = {arXiv},
  primaryClass  = {cs.CL},
  url           = {https://arxiv.org/abs/2412.07724}
}

@misc{zhao2025qwen3guard,
  author        = {Haiquan Zhao and Chenhan Yuan and Fei Huang and Xiaomeng Hu and Yichang Zhang and An Yang and Bowen Yu and Dayiheng Liu and Jingren Zhou and Junyang Lin and Baosong Yang and Chen Cheng and Jialong Tang and Jiandong Jiang and Jianwei Zhang and Jijie Xu and Ming Yan and Minmin Sun and Pei Zhang and Pengjun Xie and Qiaoyu Tang and Qin Zhu and Rong Zhang and Shibin Wu and Shuo Zhang and Tao He and Tianyi Tang and Tingyu Xia and Wei Liao and Weizhou Shen and Wenbiao Yin and Wenmeng Zhou and Wenyuan Yu and Xiaobin Wang and Xiaodong Deng and Xiaodong Xu and Xinyu Zhang and Yang Liu and Yeqiu Li and Yi Zhang and Yong Jiang and Yu Wan and Yuxin Zhou},
  title         = {{Qwen3Guard} Technical Report},
  year          = {2025},
  eprint        = {2510.14276},
  archivePrefix = {arXiv},
  primaryClass  = {cs.CL},
  url           = {https://arxiv.org/abs/2510.14276}
}

@misc{nvidia2024nemoguard,
  author       = {{NVIDIA}},
  title        = {{Llama-3.1-NemoGuard-8B-ContentSafety}},
  year         = {2024},
  howpublished = {NVIDIA NIM model card, Hugging Face, \url{https://huggingface.co/nvidia/llama-3.1-nemoguard-8b-content-safety}},
  note         = {Methodology described in Ghosh et al.\ (AEGIS2.0, 2025)}
}

@inproceedings{kumar2025polyguard,
  author        = {Priyanshu Kumar and Devansh Jain and Akhila Yerukola and Liwei Jiang and Himanshu Beniwal and Thomas Hartvigsen and Maarten Sap}, 
  title         = {{PolyGuard}: A Multilingual Safety Moderation Tool for 17 Languages},
  booktitle     = {Proceedings of the Conference on Language Modeling (COLM)},
  year          = {2025},
  eprint        = {2504.04377},
  archivePrefix = {arXiv},
  url           = {https://arxiv.org/abs/2504.04377}
}

@inproceedings{verma2025omniguard,
  author        = {Sahil Verma and Keegan Hines and Jeff Bilmes and Charlotte Siska and Luke Zettlemoyer and Hila Gonen and Chandan Singh}, 
  title         = {{MULTIGUARD}: An Efficient Approach for {AI} Safety Moderation Across Languages and Modalities},
  booktitle     = {Proceedings of the 2025 Conference on Empirical Methods in Natural Language Processing},
  year          = {2025},
  pages         = {16173--16187},
  publisher     = {Association for Computational Linguistics},
  eprint        = {2505.23856},
  archivePrefix = {arXiv},
  url           = {https://aclanthology.org/2025.emnlp-main.819/},
  doi           = {10.18653/v1/2025.emnlp-main.819}
}

@inproceedings{wang2025standguard,
  author        = {Minjia Wang and Pingping Lin and Siqi Cai and Shengnan An and Shengjie Ma and Zeqi Lin and Congrui Huang and Bixiong Xu},
  title         = {{STAND-Guard}: A Small Task-Adaptive Content Moderation Model},
  booktitle     = {Proceedings of the 31st International Conference on Computational Linguistics: Industry Track (COLING)},
  year          = {2025},
  eprint        = {2411.05214},
  archivePrefix = {arXiv},
  url           = {https://aclanthology.org/2025.coling-industry.1/}
}

@misc{joshi2025cultureguard,
  author        = {Raviraj Joshi and Rakesh Paul and Kanishk Singla and Anusha Kamath and Michael Evans and Katherine Luna and Shaona Ghosh and Utkarsh Vaidya and Eileen Long and Sanjay Singh Chauhan and Niranjan Wartikar},
  title         = {{CultureGuard}: Towards Culturally-Aware Dataset and Guard Model for Multilingual Safety Applications},
  year          = {2025},
  eprint        = {2508.01710},
  archivePrefix = {arXiv},
  primaryClass  = {cs.CL},
  url           = {https://arxiv.org/abs/2508.01710}
}

@misc{deng2025duoguard,
  author        = {Yihe Deng and Yu Yang and Junkai Zhang and Wei Wang and Bo Li},
  title         = {{DuoGuard}: A Two-Player {RL}-Driven Framework for Multilingual {LLM} Guardrails},
  year          = {2025},
  eprint        = {2502.05163},
  archivePrefix = {arXiv},
  primaryClass  = {cs.CL},
  url           = {https://arxiv.org/abs/2502.05163}
}

@inproceedings{mazeika2024harmbench,
  author        = {Mantas Mazeika and Long Phan and Xuwang Yin and Andy Zou and Zifan Wang and Norman Mu and Elham Sakhaee and Nathaniel Li and Steven Basart and Bo Li and David Forsyth and Dan Hendrycks}, 
  title         = {{HarmBench}: A Standardized Evaluation Framework for Automated Red Teaming and Robust Refusal}, 
  booktitle     = {Proceedings of the 41st International Conference on Machine Learning (ICML)},
  year          = {2024},
  eprint        = {2402.04249},
  archivePrefix = {arXiv},
  url           = {https://arxiv.org/abs/2402.04249}
}

@inproceedings{lin2023toxicchat,
  author        = {Zi Lin and Zihan Wang and Yongqi Tong and Yangkun Wang and Yuxin Guo and Yujia Wang and Jingbo Shang},
  title         = {{ToxicChat}: Unveiling Hidden Challenges of Toxicity Detection in Real-World User-{AI} Conversation},
  booktitle     = {Findings of the Association for Computational Linguistics: EMNLP 2023},
  year          = {2023},
  eprint        = {2310.17389},
  archivePrefix = {arXiv},
  url           = {https://aclanthology.org/2023.findings-emnlp.311/}
}

@inproceedings{ji2023beavertails,
  author        = {Jiaming Ji and Mickel Liu and Juntao Dai and Xuehai Pan and Chi Zhang and Ce Bian and Ruiyang Sun and Yizhou Wang and Yaodong Yang},
  title         = {{BeaverTails}: Towards Improved Safety Alignment of {LLM} via a Human-Preference Dataset}, 
  booktitle     = {Advances in Neural Information Processing Systems 36: Datasets and Benchmarks Track (NeurIPS)},
  year          = {2023},
  eprint        = {2307.04657},
  archivePrefix = {arXiv},
  url           = {https://arxiv.org/abs/2307.04657}
}

@inproceedings{rottger2024xstest,
  author        = {Paul R{\"o}ttger and Hannah Rose Kirk and Bertie Vidgen and Giuseppe Attanasio and Federico Bianchi and Dirk Hovy},
  title         = {{XSTest}: A Test Suite for Identifying Exaggerated Safety Behaviours in Large Language Models}, 
  booktitle     = {Proceedings of the 2024 Conference of the North American Chapter of the Association for Computational Linguistics (NAACL)},
  year          = {2024},
  eprint        = {2308.01263},
  archivePrefix = {arXiv},
  url           = {https://aclanthology.org/2024.naacl-long.301/}
}

@inproceedings{cui2025orbench,
  author        = {Justin Cui and Wei-Lin Chiang and Ion Stoica and Cho-Jui Hsieh},
  title         = {{OR-Bench}: An Over-Refusal Benchmark for Large Language Models},
  booktitle     = {Proceedings of the 42nd International Conference on Machine Learning (ICML)},
  year          = {2025},
  eprint        = {2405.20947},
  archivePrefix = {arXiv},
  url           = {https://arxiv.org/abs/2405.20947}
}

@inproceedings{an2024automatic,
  author        = {Bang An and Sicheng Zhu and Ruiyi Zhang and Michael-Andrei Panaitescu-Liess and Yuancheng Xu and Furong Huang},
  title         = {Automatic Pseudo-Harmful Prompt Generation for Evaluating False Refusals in Large Language Models},
  booktitle     = {First Conference on Language Modeling},
  year          = {2024},
  url           = {https://openreview.net/forum?id=ljFgX6A8NL}
}

@inproceedings{shen2024anythingnow,
  author        = {Xinyue Shen and Zeyuan Chen and Michael Backes and Yun Shen and Yang Zhang},
  title         = {``{D}o Anything Now'': Characterizing and Evaluating In-The-Wild Jailbreak Prompts on Large Language Models}, 
  booktitle     = {Proceedings of the 2024 ACM SIGSAC Conference on Computer and Communications Security (CCS)},
  year          = {2024},
  eprint        = {2308.03825},
  archivePrefix = {arXiv},
  url           = {https://arxiv.org/abs/2308.03825}
}

@misc{zou2023universal,
  author        = {Andy Zou and Zifan Wang and Nicholas Carlini and Milad Nasr and J. Zico Kolter and Matt Fredrikson},
  title         = {Universal and Transferable Adversarial Attacks on Aligned Language Models},
  year          = {2023},
  eprint        = {2307.15043},
  archivePrefix = {arXiv},
  primaryClass  = {cs.CL},
  url           = {https://arxiv.org/abs/2307.15043}
}

@inproceedings{wei2023jailbroken,
  author        = {Alexander Wei and Nika Haghtalab and Jacob Steinhardt},
  title         = {Jailbroken: How Does {LLM} Safety Training Fail?}, 
  booktitle     = {Advances in Neural Information Processing Systems 36 (NeurIPS)},
  year          = {2023},
  eprint        = {2307.02483},
  archivePrefix = {arXiv},
  url           = {https://arxiv.org/abs/2307.02483}
}

@misc{peysakhovich2023attention,
  author    = {Alexander Peysakhovich and Adam Lerer},
  title     = {Attention Sorting Combats Recency Bias In Long Context Language Models},
  year      = {2023},
  eprint    = {2310.01427},
  archivePrefix = {arXiv},
  primaryClass  = {cs.CL},
  url       = {https://arxiv.org/abs/2310.01427}
}

@inproceedings{song2021scoresde,
  author    = {Yang Song and Jascha Sohl-Dickstein and Diederik P. Kingma and Abhishek Kumar and Stefano Ermon and Ben Poole},
  title     = {Score-Based Generative Modeling through Stochastic Differential Equations},
  booktitle = {International Conference on Learning Representations (ICLR)},
  year      = {2021},
  eprint    = {2011.13456},
  archivePrefix = {arXiv}
}

@inproceedings{hoogeboom2021multinomial,
  author    = {Emiel Hoogeboom and Didrik Nielsen and Priyank Jaini and Patrick Forr{\'e} and Max Welling},
  title     = {Argmax Flows and Multinomial Diffusion: Learning Categorical Distributions},
  booktitle = {Advances in Neural Information Processing Systems 34 (NeurIPS)},
  year      = {2021},
  eprint    = {2102.05379},
  archivePrefix = {arXiv}
}

@inproceedings{austin2021d3pm,
  author    = {Jacob Austin and Daniel D. Johnson and Jonathan Ho and Daniel Tarlow and Rianne van den Berg},
  title     = {Structured Denoising Diffusion Models in Discrete State-Spaces},
  booktitle = {Advances in Neural Information Processing Systems 34 (NeurIPS)},
  year      = {2021},
  eprint    = {2107.03006},
  archivePrefix = {arXiv}
}

@inproceedings{campbell2022ctmc,
  author    = {Andrew Campbell and Joe Benton and Valentin De Bortoli and Thomas Rainforth and George Deligiannidis and Arnaud Doucet},
  title     = {A Continuous Time Framework for Discrete Denoising Models},
  booktitle = {Advances in Neural Information Processing Systems 35 (NeurIPS)},
  year      = {2022},
  eprint    = {2205.14987},
  archivePrefix = {arXiv}
}

@inproceedings{lou2024sedd,
  author    = {Aaron Lou and Chenlin Meng and Stefano Ermon},
  title     = {Discrete Diffusion Modeling by Estimating the Ratios of the Data Distribution},
  booktitle = {Proceedings of the 41st International Conference on Machine Learning (ICML)},
  year      = {2024},
  eprint    = {2310.16834},
  archivePrefix = {arXiv}
}

@inproceedings{sahoo2024mdlm,
  author    = {Subham Sekhar Sahoo and Marianne Arriola and Yair Schiff and Aaron Gokaslan and Edgar Marroquin and Justin T. Chiu and Alexander Rush and Volodymyr Kuleshov},
  title     = {Simple and Effective Masked Diffusion Language Models},
  booktitle = {Advances in Neural Information Processing Systems 37 (NeurIPS)},
  year      = {2024},
  eprint    = {2406.07524},
  archivePrefix = {arXiv}
}

@inproceedings{li2023diffusion_classifier,
  author    = {Alexander C. Li and Mihir Prabhudesai and Shivam Duggal and Ellis Brown and Deepak Pathak},
  title     = {Your Diffusion Model is Secretly a Zero-Shot Classifier},
  booktitle = {Proceedings of the IEEE/CVF International Conference on Computer Vision (ICCV)},
  pages     = {2206--2217},
  year      = {2023}
}

@inproceedings{clark2023text,
  author    = {Kevin Clark and Priyank Jaini},
  title     = {Text-to-Image Diffusion Models are Zero-Shot Classifiers},
  booktitle = {Advances in Neural Information Processing Systems 36 (NeurIPS)},
  year      = {2023}
}

@misc{yogatama2017generative,
  author    = {Dani Yogatama and Chris Dyer and Wang Ling and Phil Blunsom},
  title     = {Generative and Discriminative Text Classification with Recurrent Neural Networks},
  year      = {2017},
  eprint    = {1703.01898},
  archivePrefix = {arXiv},
  primaryClass  = {stat.ML},
  url       = {https://arxiv.org/abs/1703.01898}
}

@article{geirhos2020shortcut,
  author  = {Robert Geirhos and J{\"o}rn-Henrik Jacobsen and Claudio Michaelis and Richard Zemel and Wieland Brendel and Matthias Bethge and Felix A. Wichmann},
  title   = {Shortcut Learning in Deep Neural Networks},
  journal = {Nature Machine Intelligence},
  volume  = {2},
  number  = {11},
  pages   = {665--673},
  year    = {2020},
  doi     = {10.1038/s42256-020-00257-z}
}

@misc{weidinger2021ethical,
  author        = {Laura Weidinger and John Mellor and Maribeth Rauh and Conor Griffin and Jonathan Uesato and Po-Sen Huang and Myra Cheng and Mia Glaese and Borja Balle and Atoosa Kasirzadeh and Zac Kenton and Sasha Brown and Will Hawkins and Tom Stepleton and Courtney Biles and Abeba Birhane and Julia Haas and Laura Rimell and Lisa Anne Hendricks and William Isaac and Sean Legassick and Geoffrey Irving and Iason Gabriel},
  title         = {Ethical and Social Risks of Harm from Language Models},
  year          = {2021},
  eprint        = {2112.04359},
  archivePrefix = {arXiv},
  primaryClass  = {cs.CL},
  url           = {https://arxiv.org/abs/2112.04359}
}

@misc{hendrycks2023overview,
  author        = {Dan Hendrycks and Mantas Mazeika and Thomas Woodside},
  title         = {An Overview of Catastrophic {AI} Risks},
  year          = {2023},
  eprint        = {2306.12001},
  archivePrefix = {arXiv},
  primaryClass  = {cs.CY},
  url           = {https://arxiv.org/abs/2306.12001}
}

@article{anwar2024foundational,
  author        = {Usman Anwar and Abulhair Saparov and Javier Rando and Daniel Paleka and Miles Turpin and Peter Hase and Ekdeep Singh Lubana and Erik Jenner and Stephen Casper and Oliver Sourbut and Benjamin L. Edelman and Zhaowei Zhang and Mario G{\"u}nther and Anton Korinek and Jose Hernandez-Orallo and Lewis Hammond and Eric Bigelow and Alexander Pan and Lauro Langosco and Tomasz Korbak and Heidi Zhang and Ruiqi Zhong and Se{\'a}n {\'O} h{\'E}igeartaigh and Gabriel Recchia and Giulio Corsi and Alan Chan and Markus Anderljung and Lilian Edwards and Aleksandar Petrov and Christian Schroeder de Witt and Sumeet Ramesh Motwani and Yoshua Bengio and Danqi Chen and Philip H. S. Torr and Samuel Albanie and Tegan Maharaj and Jakob Foerster and Florian Tram{\`e}r and He He and Atoosa Kasirzadeh and Yejin Choi and David Krueger},
  title         = {Foundational Challenges in Assuring Alignment and Safety of Large Language Models},
  journal       = {Transactions on Machine Learning Research},
  volume        = {2024},
  year          = {2024},
  eprint        = {2404.09932},
  archivePrefix = {arXiv},
  primaryClass  = {cs.LG},
  url           = {https://openreview.net/forum?id=oVTkOs8Pka}
}

@inproceedings{zheng2024lmsys,
  author    = {Lianmin Zheng and Wei-Lin Chiang and Ying Sheng and Tianle Li and Siyuan Zhuang and Zhanghao Wu and Yonghao Zhuang and Zhuohan Li and Zi Lin and Eric P. Xing and Joseph E. Gonzalez and Ion Stoica and Hao Zhang},
  title     = {{LMSYS-Chat-1M}: A Large-Scale Real-World {LLM} Conversation Dataset},
  booktitle = {International Conference on Learning Representations (ICLR)},
  year      = {2024},
  eprint    = {2309.11998},
  archivePrefix = {arXiv},
  url       = {https://arxiv.org/abs/2309.11998}
}

@inproceedings{zhao2024wildchat,
  author    = {Wenting Zhao and Xiang Ren and Jack Hessel and Claire Cardie and Yejin Choi and Yuntian Deng},
  title     = {{WildChat}: {1M} {ChatGPT} Interaction Logs in the Wild},
  booktitle = {International Conference on Learning Representations (ICLR)},
  year      = {2024},
  eprint    = {2405.01470},
  archivePrefix = {arXiv},
  url       = {https://arxiv.org/abs/2405.01470}
}

@inproceedings{dai2024saferlhf,
  author    = {Josef Dai and Xuehai Pan and Ruiyang Sun and Jiaming Ji and Xinbo Xu and Mickel Liu and Yizhou Wang and Yaodong Yang},
  title     = {{Safe RLHF}: Safe Reinforcement Learning from Human Feedback},
  booktitle = {International Conference on Learning Representations (ICLR)},
  year      = {2024},
  eprint    = {2310.12773},
  archivePrefix = {arXiv},
  url       = {https://arxiv.org/abs/2310.12773}
}

@article{llm-assistant-healthcare,
  author       = {Ziqi Yang and
                  Xuhai Xu and
                  Bingsheng Yao and
                  Ethan Rogers and
                  Shao Zhang and
                  Stephen S. Intille and
                  Nawar Shara and
                  Guodong Gordon Gao and
                  Dakuo Wang},
  title        = {Talk2Care: An LLM-based Voice Assistant for Communication between
                  Healthcare Providers and Older Adults},
  journal      = {Proc. {ACM} Interact. Mob. Wearable Ubiquitous Technol.},
  volume       = {8},
  number       = {2},
  pages        = {73:1--73:35},
  year         = {2024},
  doi          = {10.1145/3659625}
}

@article{llm-software-engineering,
  author       = {Junda He and
                  Christoph Treude and
                  David Lo},
  title        = {LLM-Based Multi-Agent Systems for Software Engineering: Literature
                  Review, Vision, and the Road Ahead},
  journal      = {{ACM} Trans. Softw. Eng. Methodol.},
  volume       = {34},
  number       = {5},
  pages        = {124:1--124:30},
  year         = {2025},
  doi          = {10.1145/3712003}
}

@inproceedings{llm-education,
  author       = {Zhendong Chu and
                  Shen Wang and
                  Jian Xie and
                  Tinghui Zhu and
                  Yibo Yan and
                  Jinheng Ye and
                  Aoxiao Zhong and
                  Xuming Hu and
                  Jing Liang and
                  Philip S. Yu and
                  Qingsong Wen},
  title        = {{LLM} Agents for Education: Advances and Applications},
  booktitle    = {Findings of the Association for Computational Linguistics: EMNLP 2025},
  pages        = {13782--13810},
  publisher    = {Association for Computational Linguistics},
  year         = {2025},
  eprint       = {2503.11733},
  archivePrefix = {arXiv},
  url          = {https://aclanthology.org/2025.findings-emnlp.743/},
  doi          = {10.18653/v1/2025.findings-emnlp.743}
}

@inproceedings{openai-mod,
  author       = {Todor Markov and
                  Chong Zhang and
                  Sandhini Agarwal and
                  Florentine Eloundou Nekoul and
                  Theodore Lee and
                  Steven Adler and
                  Angela Jiang and
                  Lilian Weng},
  title        = {A Holistic Approach to Undesired Content Detection in the Real World},
  booktitle    = {Proceedings of the AAAI Conference on Artificial Intelligence},
  volume       = {37},
  number       = {12},
  pages        = {15009--15018},
  publisher    = {{AAAI} Press},
  year         = {2023},
  doi          = {10.1609/aaai.v37i12.26752}
}

@inproceedings{piguard-notinject,
  author       = {Hao Li and
                  Xiaogeng Liu and
                  Ning Zhang and
                  Chaowei Xiao},
  title        = {PIGuard: Prompt Injection Guardrail via Mitigating Overdefense for
                  Free},
  booktitle    = {Proceedings of the 63rd Annual Meeting of the Association for Computational Linguistics (Volume 1: Long Papers)},
  pages        = {30420--30437},
  publisher    = {Association for Computational Linguistics},
  year         = {2025},
  url          = {https://aclanthology.org/2025.acl-long.1468/},
  doi          = {10.18653/v1/2025.acl-long.1468}
}
%%%%%%%%%%%%%%%%%%%%%%%%%%%%%%%%%%%%%%%%%%%%%%%%%%%%%%%%%%%%

\appendix

\crefalias{section}{appendix}
\crefalias{subsection}{appendix}

\section{Dataset Preprocessing and Model Usage}
\label{app:datasets-models}

For prompt-only moderation, we evaluate OpenAI Moderation~\citep{openai-mod}, ToxicChat~\citep{lin2023toxicchat}, and XSTest~\citep{rottger2024xstest}. For prompt-response moderation, we evaluate BeaverTails~\citep{ji2023beavertails}, PKU-SafeRLHF~\citep{dai2024saferlhf}, and XSTest-Response from WildGuard~\citep{han2024wildguardopenonestopmoderation}. All datasets are mapped to binary \texttt{safe}/\texttt{unsafe} labels, with any dataset-provided harmful, toxic, policy-violating, or refusal-requiring label mapped to \texttt{unsafe}.

In our evaluation of SafeRLHF, we follow \citet{han2024wildguardopenonestopmoderation} and retain only items whose prompt is paired with both a safe and an unsafe response, while subsampling 1K prompt--response pairs to reduce evaluation cost. For PHTest~\citep{an2024automatic}, we evaluate only on samples annotated as ``harmless'' rather than controversial, resulting in a dataset version where all examples are \texttt{safe}. OpenAI Moderation, ToxicChat, XSTest, BeaverTails, and XSTest-Response are otherwise used as provided by their released benchmark splits.

\textbf{Baseline models.}
We evaluate the released Hugging Face checkpoints for ShieldGemma 2B/9B/27B~\citep{zeng2024shieldgemma}, Granite Guardian 3.1 2B and 3.2 5B~\citep{padhi2024graniteguardian}, Llama Guard 3 8B~\citep{metallamaguard3}, WildGuard 7B~\citep{han2024wildguardopenonestopmoderation}, NemoGuard 8B~\citep{nvidia2024nemoguard}, Llama Guard 4 12B~\citep{metallamaguard4}, and Qwen3Guard-Gen 0.6B/4B/8B~\citep{zhao2025qwen3guard}. We use each model's official Hugging Face chat template via \texttt{apply\_chat\_template} and the recommended inference settings.
For Qwen3Guard we use the ``strict'' mode as it is the highest performing variant.

\section{Additional Held-Out-Benchmark Results}
\label{app:ood-calibration}

\subsection{Direct-verdict ablation}
\label{app:direct-verdict-ablation}

For the predictive performance results below, we additionally consider \textbf{\algo Direct} as an ablation.
It uses the same model weights as \algo, but simulates the standard guard-model paradigm.
Instead of scoring the moderated region under competing class-conditional hypotheses, \algo Direct only predicts the probabilities for the verdict (label) token $y \in \{\texttt{safe}, \texttt{unsafe}\}$ placed after the prompt and, when applicable, the response.
In the masked-diffusion model, prediction equates to unmasking $y$. This setting, therefore, isolates the effect of the proposed class-conditional reconstruction objective from that of using LLaDA as the underlying model.

\begin{table*}[h]
\centering
\small
\setlength{\tabcolsep}{4pt}
\caption{Direct-verdict ablation on held-out AUPRC. $\Delta$ is \algo minus \algo Direct.}
\label{tab:llada-direct-auprc-ablation}
\resizebox{\textwidth}{!}{%
\begin{tabular}{lcccccc}
\toprule
Model & ToxicChat & OpenAI Mod & XSTest & BeaverTails & XSTest-Resp & Avg \\
\midrule
LLaDA-Guard Direct 8B & 0.8163 & 0.8163 & 0.9594 & 0.9104 & 0.9412 & 0.8887 \\
LLaDA-Guard 8B & \textbf{0.8452} & \textbf{0.8568} & \textbf{0.9691} & \textbf{0.9294} & \textbf{0.9806} & \textbf{0.9162} \\
\midrule
$\Delta$ & +0.0289 & +0.0405 & +0.0097 & +0.0190 & +0.0394 & +0.0275 \\
\bottomrule
\end{tabular}%
}
\end{table*}

\begin{table*}[h]
\centering
\small
\setlength{\tabcolsep}{4pt}
\caption{Direct-verdict ablation on held-out F1 at the default decision threshold used in evaluation. $\Delta$ is \algo minus \algo Direct.}
\label{tab:llada-direct-f1-ablation}
\resizebox{\textwidth}{!}{%
\begin{tabular}{lccccccc}
\toprule
Model & ToxicChat & OpenAI Mod & XSTest & BeaverTails & XSTest-Resp & Avg \\
\midrule
LLaDA-Guard Direct 8B & 0.7566 & 0.7339 & 0.8855 & 0.7342 & 0.8514 & 0.7923 \\
LLaDA-Guard 8B & \textbf{0.7755} & \textbf{0.7513} & \textbf{0.8877}  & \textbf{0.8329} & \textbf{0.9281} & \textbf{0.8599} \\
\midrule
$\Delta$ & +0.0189 & +0.0174 & +0.0022 & +0.0877 & +0.0767 & +0.0406 \\
\bottomrule
\end{tabular}%
}
\end{table*}

\Cref{tab:llada-direct-auprc-ablation,tab:llada-direct-f1-ablation} show that class-conditional reconstruction improves over direct verdict-token scoring with the same LLaDA weights on every held-out dataset. The largest fixed-threshold gains occur on response datasets, where the average F1 improves by $4.06$ percentage points over \algo Direct.

\subsection{Calibration breakdowns}

\begin{table}[h]
\centering
\scriptsize
\setlength{\tabcolsep}{4pt}
\caption{Held-out-benchmark calibration summary ($\times 100$, lower is better). Avg.\ over ToxicChat, OpenAI Mod, XSTest, SafeRLHF, BeaverTails, XSTest-Response, PHTest Harmless.}
\label{tab:heldout-calibration-summary}
\begin{tabular}{lrrr}
\toprule
Model & ECE & Brier & NLL \\
\midrule
ShieldGemma 2B          & 19.79 & 18.26 &  64.21 \\
ShieldGemma 9B          & 11.90 & 11.85 &  40.35 \\
ShieldGemma 27B         & 16.05 & 16.60 &  50.78 \\
Llama-Guard 3 8B        & \underline{9.51} & 9.84 & 47.37 \\
Llama-Guard 4 12B       & 12.74 & 12.65 & 168.65 \\
Granite Guardian 3.0 2B & 25.77 & 20.71 &  62.58 \\
Granite Guardian 3.2 5B & 11.81 & 10.81 & \underline{35.28} \\
Qwen3Guard 0.6B         & 16.61 & 13.64 &  51.57 \\
Qwen3Guard 4B           & 14.21 & 11.77 &  48.71 \\
Qwen3Guard 8B           & 13.84 & 11.29 &  48.33 \\
NemoGuard 8B            & 12.63 & 10.40 &  50.17 \\
WildGuard 7B            & 10.91 & 10.37 &  47.86 \\
\midrule
LLaDA-Guard 8B          & \textbf{8.75} & \textbf{8.45} & \textbf{32.56} \\
\bottomrule
\end{tabular}
\end{table}

\begin{table*}[h]
\centering
\small
\setlength{\tabcolsep}{4pt}
\caption{Held-out-benchmark expected calibration error (ECE) comparison. Avg is the mean over the displayed datasets in this table.}
\label{tab:full-comparison-ood-ece}
\resizebox{\textwidth}{!}{%
\begin{tabular}{lcccccccc}
\toprule
Model & ToxicChat & OpenAI Mod & XSTest & SafeRLHF & BeaverTails & XSTest-Resp & PHTest & Avg \\
\midrule
ShieldGemma 27B & 0.0419 & 0.0611 & 0.2081 & 0.2936 & 0.3553 & 0.0424 & 0.1214 & 0.1605 \\
ShieldGemma 2B & 0.0950 & 0.2106 & 0.1568 & 0.3540 & 0.4236 & 0.0812 & 0.0639 & 0.1979 \\
ShieldGemma 9B & 0.0252 & 0.0878 & 0.1065 & 0.2087 & 0.2105 & 0.0603 & 0.1341 & 0.1190 \\
Llama-Guard 4 12B & 0.1179 & 0.1546 & 0.1331 & 0.1121 & 0.2707 & 0.0337 & 0.0697 & 0.1274 \\
Llama-Guard 3 8B & 0.0773 & 0.0705 & 0.0774 & 0.0741 & 0.2664 & 0.0248 & 0.0753 & 0.0951 \\
Granite Guardian 3.0 2B & 0.3849 & 0.2734 & 0.2284 & 0.0542 & 0.1762 & 0.0331 & 0.6539 & 0.2577 \\
Granite Guardian 3.2 5B & 0.0535 & 0.1654 & 0.1188 & 0.0429 & 0.1269 & 0.0245 & 0.2947 & 0.1181 \\
NemoGuard 8B & 0.0487 & 0.1676 & 0.1046 & 0.0383 & 0.1854 & 0.0346 & 0.3046 & 0.1263 \\
Qwen3Guard 0.6B & 0.1546 & 0.2880 & 0.1203 & 0.1168 & 0.1052 & 0.0332 & 0.3444 & 0.1661 \\
Qwen3Guard 4B & 0.1167 & 0.2696 & 0.0804 & 0.1117 & 0.1193 & 0.0336 & 0.2633 & 0.1421 \\
Qwen3Guard 8B & 0.1199 & 0.2688 & 0.0790 & 0.1142 & 0.1199 & 0.0320 & 0.2354 & 0.1384 \\
WildGuard 7B & 0.0799 & 0.2055 & 0.0325 & 0.0603 & 0.1270 & 0.0206 & 0.2382 & 0.1091 \\
LLaDA-Guard 8B & 0.0397 & 0.1601 & 0.0534 & 0.0482 & 0.1418 & 0.0190 & 0.1506 & 0.0875 \\
\bottomrule
\end{tabular}%
}
\end{table*}

\Cref{tab:heldout-calibration-summary,tab:full-comparison-ood-ece} report the corresponding calibration results. \Cref{tab:heldout-calibration-summary} summarizes average ECE, Brier score, and NLL across the held-out benchmarks, with all values multiplied by 100; lower values indicate better calibrated probabilities.
\algo obtains the best score on all three metrics, indicating that its strong classification performance does not come at the cost of unreliable confidence estimates.
\Cref{tab:full-comparison-ood-ece} breaks down ECE by dataset, showing that our calibration advantage is stable across individual benchmarks: \algo achieves the lowest average ECE overall and remains consistently competitive across prompt-only, response, and harmless-prompt settings.
Together, these results support the main claim that class-conditional reconstruction yields guard scores that are both competitive for detection and substantially better calibrated than the baselines.

% \clearpage

\section{Additional Diagnostic Failure-Mode Results}
\label{app:diagnostic-failure-modes}

This appendix collects the expanded analyses behind \Cref{sec:eval:diagnostics}.

\begin{figure}[h]
    \centering
    \includegraphics[width=\linewidth]{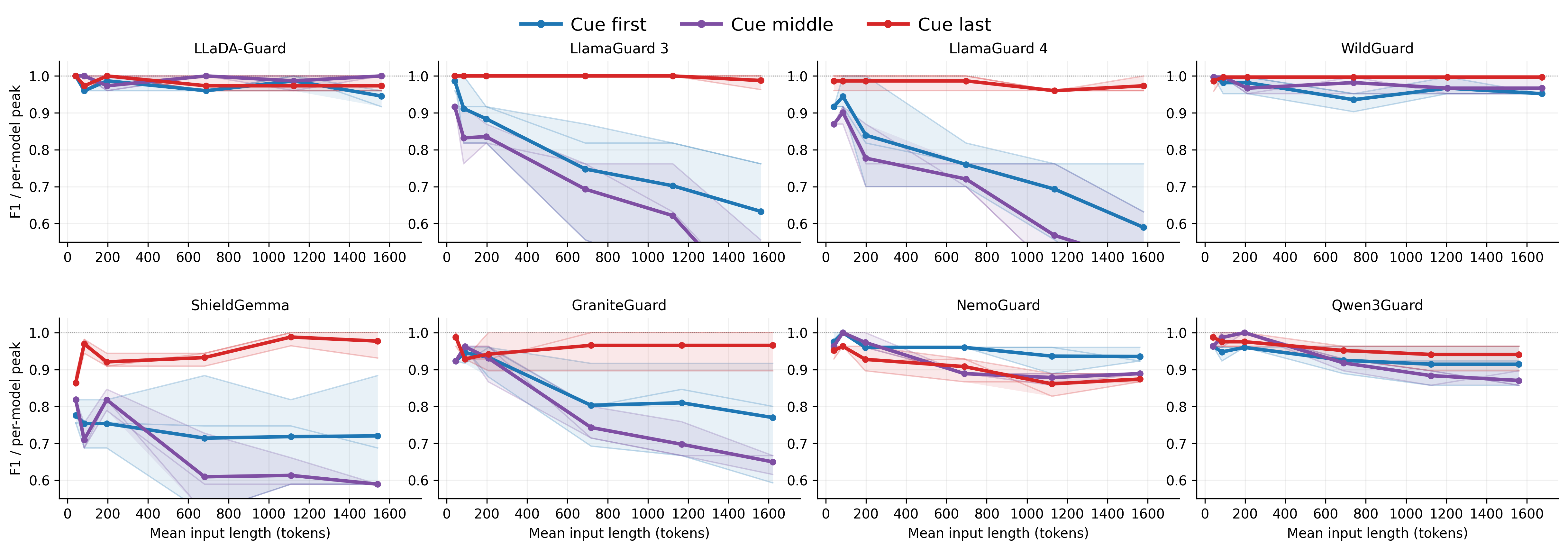}
    \caption{Per-model positional bias investigation. Each panel shows F1 normalized by that model's peak F1 across input-length buckets, with curves grouped by the position of the decisive safety cue: first, middle, or last. Flat curves as filler length increases indicate less positional sensitivity.}
    \label{fig:app-positional-sensitivity-panels}
\end{figure}

\paragraph{Positional sensitivity.}
\Cref{fig:app-positional-sensitivity-panels} expands the aggregate positional-sensitivity result from \Cref{fig:positional-sensitivity} into per-model curves. The y-axis normalizes each model by its own peak F1, so the relevant signal is the drop from the model's best operating point as filler length grows. 

\begin{figure}[h]
    \centering
    \includegraphics[width=\linewidth]{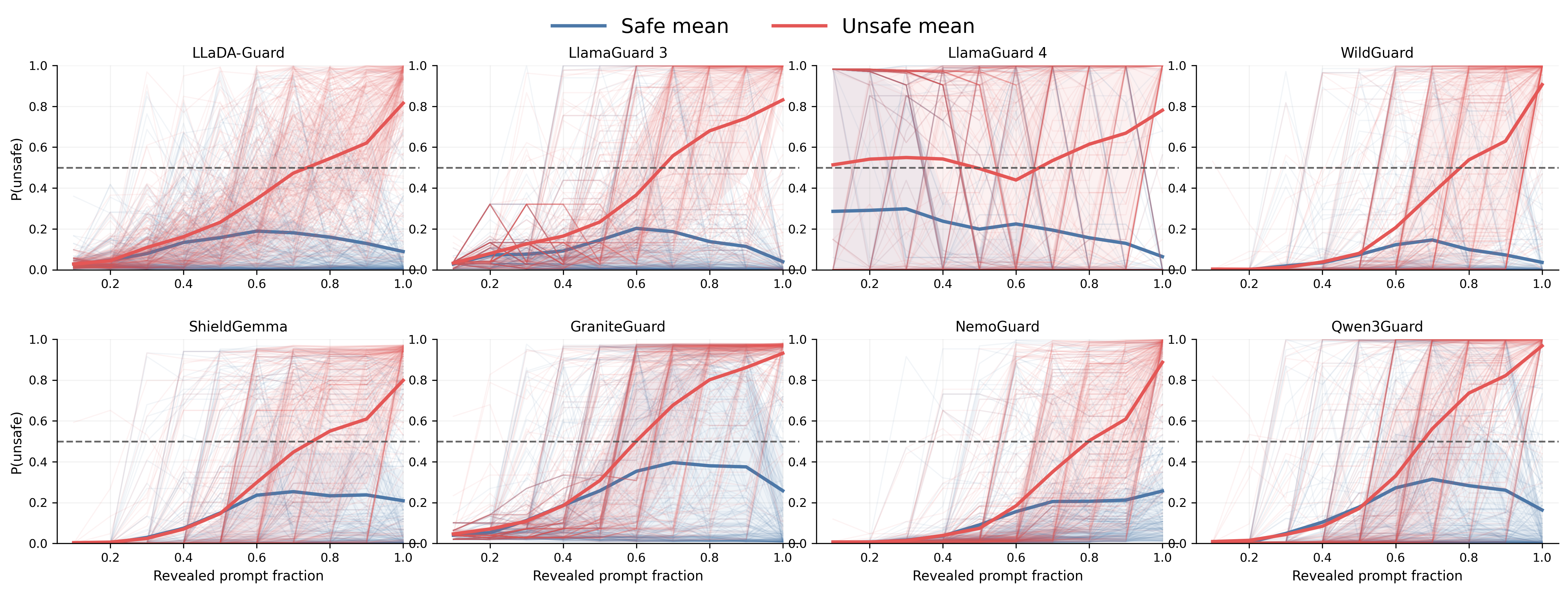}
    \caption{Prefix-score trajectories on XSTest. Each prompt is revealed from left to right, and the guard score is recomputed after each prefix. Mean \texttt{safe} and \texttt{unsafe} trajectories summarize how quickly each model crosses the \texttt{unsafe} decision boundary.}
    \label{fig:xstest_all_prefix}
\end{figure}

\paragraph{Prefix-score boundary.}
\Cref{fig:xstest_all_prefix} shows the full XSTest prefix-reveal experiment for all baselines. Each translucent line is one example and each thick line is the class mean. 

\begin{figure}[h]
    \centering
    \includegraphics[width=\linewidth]{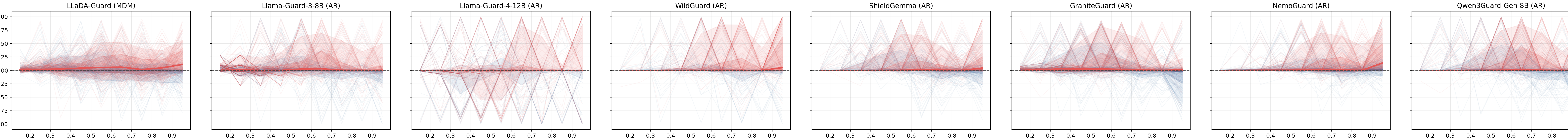}
    \caption{Prefix-score jump analysis on XSTest. Curves show changes in \texttt{unsafe} probability between consecutive revealed-prefix steps, highlighting abrupt decision-boundary jumps.}
    \label{fig:xstest_jumps}
\end{figure}

\paragraph{Prefix-score jumps.}
\Cref{fig:xstest_jumps} makes the previous observation explicit by plotting the change in \texttt{unsafe} probability between consecutive prefix steps. Large positive spikes mean that a short newly revealed fragment caused a large verdict shift.

\paragraph{Response-objective fidelity.}
\Cref{tab:misaligned-response-fidelity} tests whether response moderation is actually grounded in the response. The response is held fixed and \texttt{safe}; only the prompt changes from \texttt{safe} to \texttt{unsafe}. A response-grounded model should therefore produce nearly identical verdicts in both conditions. 
\begin{table}[h]
\centering
\small
\begin{tabular}{lcc}
\toprule
Model & Matched $\Delta P_{\texttt{unsafe}} \downarrow$ & Matched $\Delta \hat{y}_{\texttt{unsafe}} \downarrow$ \\
\midrule
  LLaDA-Guard & \textbf{0.099} & \textbf{0.060} \\
  LlamaGuard 3 & 0.206 & 0.208 \\
  LlamaGuard 4 & 0.138 & 0.132 \\
  WildGuard & 0.292 & 0.308 \\
  GraniteGuard & 0.122 & 0.140 \\
  Qwen3Guard & 0.386 & 0.424 \\
\midrule
  ShieldGemma & 0.008 & 0.000 \\
  NemoGuard & 0.019 & 0.004 \\
\bottomrule
\end{tabular}
\caption{Response-objective fidelity on matched \texttt{safe}-response pairs ($n=250$). For each pair, the same \texttt{safe} response is attached once to a \texttt{safe} prompt and once to an \texttt{unsafe} prompt, and the model is asked to classify the response. We report the increase in \texttt{unsafe} probability ($\Delta P_{\texttt{unsafe}}$) and \texttt{unsafe} prediction rate ($\Delta \hat{y}_{\texttt{unsafe}}$) when only the prompt changes from \texttt{safe} to \texttt{unsafe}. Lower is better. Rows below the rule (ShieldGemma, NemoGuard) use a separate template for prompt and response classification, which sidesteps prompt leakage by construction and yields $\Delta \approx 0$.}
\label{tab:misaligned-response-fidelity}
\end{table}

\section{NotInject Over-Defense Analysis}
\label{app:notinject-overdefense}

NotInject contains benign prompts with prompt-injection-like trigger patterns, so any \texttt{unsafe} prediction is a false positive. \Cref{tab:notinject-splits} breaks these false positives down by the number of trigger words in the prompt. \algo produces no false positives across all trigger-count splits, matching ShieldGemma and avoiding the trigger-count sensitivity seen in several strong classifiers. Granite is the most affected baseline, rising from $1.77\%$ false positives with one trigger to $10.62\%$ with three triggers; Qwen3Guard and WildGuard also become more defensive as trigger count increases. This supports the main-text over-refusal result: \algo does not achieve strong held-out-benchmark performance by simply flagging benign prompts that contain suspicious safety or injection cues.

\begin{table}[h]
\centering
\scriptsize
\setlength{\tabcolsep}{2.4pt}
\renewcommand{\arraystretch}{0.86}
\caption{NotInject over-defense by trigger-count split. All examples are benign; entries show false-positive percentage with false-positive count in parentheses, plus average false-positive confidence when any false positives occur. }
\label{tab:notinject-splits}
\begin{tabular}{l|ccc}
\toprule
Model & 1 trigger & 2 triggers & 3 triggers \\
\midrule
\algo & \begin{tabular}{@{}c@{}}0.00\% (0)\\{\tiny conf. --}\end{tabular} & \begin{tabular}{@{}c@{}}0.00\% (0)\\{\tiny conf. --}\end{tabular} & \begin{tabular}{@{}c@{}}0.00\% (0)\\{\tiny conf. --}\end{tabular} \\
Qwen3Guard & \begin{tabular}{@{}c@{}}0.00\% (0)\\{\tiny conf. --}\end{tabular} & \cellcolor{red!38}\begin{tabular}{@{}c@{}}0.88\% (1)\\{\tiny conf. 80.59\%}\end{tabular} & \cellcolor{red!58}\begin{tabular}{@{}c@{}}3.54\% (4)\\{\tiny conf. 77.55\%}\end{tabular} \\
WildGuard & \cellcolor{red!38}\begin{tabular}{@{}c@{}}0.88\% (1)\\{\tiny conf. 65.14\%}\end{tabular} & \cellcolor{red!38}\begin{tabular}{@{}c@{}}0.88\% (1)\\{\tiny conf. 99.44\%}\end{tabular} & \cellcolor{red!53}\begin{tabular}{@{}c@{}}2.65\% (3)\\{\tiny conf. 78.16\%}\end{tabular} \\
Granite & \cellcolor{red!47}\begin{tabular}{@{}c@{}}1.77\% (2)\\{\tiny conf. 60.61\%}\end{tabular} & \cellcolor{red!67}\begin{tabular}{@{}c@{}}5.31\% (6)\\{\tiny conf. 64.70\%}\end{tabular} & \cellcolor{red!88}\begin{tabular}{@{}c@{}}10.62\% (12)\\{\tiny conf. 70.61\%}\end{tabular} \\
NemoGuard & \begin{tabular}{@{}c@{}}0.00\% (0)\\{\tiny conf. --}\end{tabular} & \begin{tabular}{@{}c@{}}0.00\% (0)\\{\tiny conf. --}\end{tabular} & \cellcolor{red!47}\begin{tabular}{@{}c@{}}1.77\% (2)\\{\tiny conf. 63.69\%}\end{tabular} \\
Llama-Guard 3 & \begin{tabular}{@{}c@{}}0.00\% (0)\\{\tiny conf. --}\end{tabular} & \begin{tabular}{@{}c@{}}0.00\% (0)\\{\tiny conf. --}\end{tabular} & \cellcolor{red!38}\begin{tabular}{@{}c@{}}0.88\% (1)\\{\tiny conf. 90.47\%}\end{tabular} \\
ShieldGemma & \begin{tabular}{@{}c@{}}0.00\% (0)\\{\tiny conf. --}\end{tabular} & \begin{tabular}{@{}c@{}}0.00\% (0)\\{\tiny conf. --}\end{tabular} & \begin{tabular}{@{}c@{}}0.00\% (0)\\{\tiny conf. --}\end{tabular} \\
Llama-Guard 4 & \begin{tabular}{@{}c@{}}0.00\% (0)\\{\tiny conf. --}\end{tabular} & \begin{tabular}{@{}c@{}}0.00\% (0)\\{\tiny conf. --}\end{tabular} & \cellcolor{red!53}\begin{tabular}{@{}c@{}}2.65\% (3)\\{\tiny conf. 89.94\%}\end{tabular} \\
\bottomrule
\end{tabular}
\end{table}

\section{Aegis Validation Time-Step Sweep}
\label{app:aegis-t-sweep}

\begin{figure}[tb]
    \centering
    \includegraphics[width=0.85\linewidth]{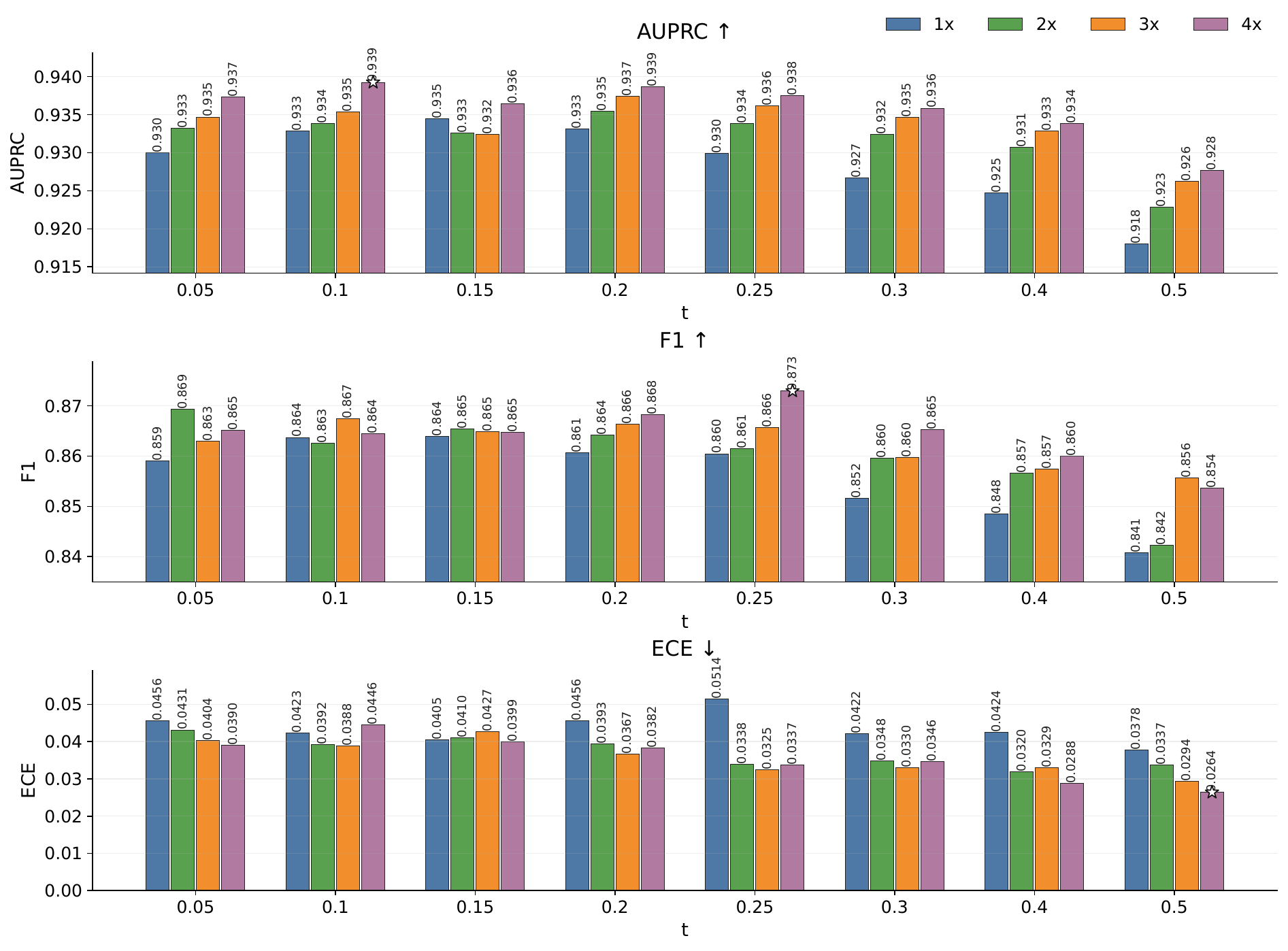}
    \caption{Aegis~2.0 validation performance across diffusion time-step values $t$ and mask sampling repetitions $K$.}
    \label{fig:aegis-val-t-sweep}
\end{figure}

\Cref{fig:aegis-val-t-sweep} evaluates LLaDA-Guard on the Aegis validation set across different diffusion time-step values $t$ and repeated mask samples. The main takeaway is that performance is robust across a broad range of masking levels: AUPRC stays near $0.93$ for $t \leq 0.4$, and F1 remains within roughly three points across the sweep. Repeats help most for calibration, with ECE consistently improving from one pass to multiple passes, offering a way to use test-time computation to boost calibration.

\section{Localization and Rewriting Details}
\label{app:localization-rewriting}

For the attention localizer, we rank tokens by their attention contribution to the verdict position. This provides a common attribution proxy for all guard models, but it is indirect: attention indicates where the verdict representation attends, not which tokens causally change the safety score.

For the leave-one-out localizer, we exploit \algo's masked scoring. For each token $i$ in the moderated region, we temporarily mask that token, recompute the safety score, and rank tokens by
\[
\Delta_i=\hat{s}_\theta(M)-\hat{s}^{(-i)}_\theta(M),
\]
where $\hat{s}^{(-i)}_\theta(M)$ denotes the score after masking token $i$. Larger positive values indicate that token $i$ contributes more strongly to the \texttt{unsafe} likelihood-ratio score. This gives a more direct attribution signal, at the cost of additional scoring passes.

For rewriting, the selected tokens are masked and infilled by \algo under the \texttt{safe} label condition. We repeat this process until the judge classifies the prompt as \texttt{safe} or the masking budget is exhausted.

\subsection{Full rewriting results}
\label{app:full-rewriting-results}

\Cref{tab:rewrite-budget-cumulative,tab:rewrite-localizer-conversion} report the full rewriting results per dataset and budget, corresponding to the main-text rewriting figure.

\begin{table}[t]
\centering
\small
\begin{tabular}{lrrrrr}
\toprule
Localizer & B1 & B2 & B3 & B4 & B5 \\
\midrule
ShieldGemma 27B & 10.9 & 20.9 & 29.8 & 38.8 & 43.4 \\
ShieldGemma 2B & 16.5 & 32.9 & 41.4 & 47.3 & 50.7 \\
ShieldGemma 9B & 12.2 & 21.0 & 27.5 & 32.9 & 40.9 \\
Llama-Guard 4 12B & 14.7 & 26.1 & 33.5 & 39.3 & 44.7 \\
Llama-Guard 3 8B & 12.4 & 23.0 & 32.6 & 41.7 & 48.6 \\
Granite Guardian 2B & 13.7 & 24.8 & 34.3 & 40.6 & 45.9 \\
Granite Guardian 5B & 19.6 & 34.7 & 43.1 & 47.2 & 50.1 \\
NemoGuard 8B & 12.0 & 27.2 & 40.1 & 46.4 & 51.0 \\
Qwen3Guard 0.6B & 13.7 & 22.3 & 31.5 & 37.2 & 42.7 \\
Qwen3Guard 4B & 14.3 & 28.4 & 36.3 & 43.3 & 47.4 \\
Qwen3Guard 8B & 14.1 & 27.4 & 36.5 & 43.3 & 48.1 \\
WildGuard 7B & 14.0 & 32.9 & 42.4 & 48.9 & 52.1 \\
LLaDA-Guard 8B & 21.3 & 38.0 & 46.4 & 50.7 & 53.0 \\
LLaDA-Guard 8B Masked & 37.6 & 47.9 & 53.3 & 57.1 & 60.7 \\
\bottomrule
\end{tabular}
\caption{Dataset-balanced cumulative conversion-to-safe rate (\%) by rewrite budget. B1--B5 denote increasing token-rewrite budgets; each value averages over XSTest, ToxicChat, and OpenAI Mod.}
\label{tab:rewrite-budget-cumulative}
\end{table}

\begin{table}[t]
\centering
\small
\begin{tabular}{lcccc}
\toprule
Localizer & XSTest & ToxicChat & OpenAI Mod & Avg \\
\midrule
ShieldGemma 27B & 85.0 & 35.1 & 10.1 & 43.4 \\
ShieldGemma 2B & 96.0 & 42.3 & 13.8 & 50.7 \\
ShieldGemma 9B & 77.0 & 34.5 & 11.1 & 40.9 \\
Llama-Guard 4 12B & 85.5 & 37.0 & 11.5 & 44.7 \\
Llama-Guard 3 8B & 92.0 & 41.4 & 12.4 & 48.6 \\
Granite Guardian 2B & 91.0 & 37.0 & 9.7 & 45.9 \\
Granite Guardian 5B & 92.5 & 43.9 & 14.0 & 50.1 \\
NemoGuard 8B & 95.0 & 45.3 & 12.8 & 51.0 \\
Qwen3Guard 0.6B & 84.5 & 35.4 & 8.2 & 42.7 \\
Qwen3Guard 4B & 93.5 & 37.6 & 11.3 & 47.4 \\
Qwen3Guard 8B & 96.0 & 38.1 & 10.1 & 48.1 \\
WildGuard 7B & 98.0 & 45.6 & 12.8 & 52.1 \\
LLaDA-Guard 8B & 99.0 & 44.8 & 15.1 & 53.0 \\
LLaDA-Guard 8B Masked & 99.0 & 55.0 & 17.0 & 60.7 \\
\bottomrule
\end{tabular}
\caption{Conversion-to-safe rate (\%) by localizer and dataset at a rewrite budget of 5.}
\label{tab:rewrite-localizer-conversion}
\end{table}

\FloatBarrier

\section{Limitations, Broader Impact, and Compute}
\label{app:limitations-impact-compute}

\subsection{Limitations}
\label{app:limitations}

Our results are limited by the scope of the evaluated datasets and policies. The main evaluation focuses on English prompt and response moderation benchmarks with binary \texttt{safe}/\texttt{unsafe} labels; we do not evaluate multilingual, multimodal, or policy-customized moderation settings. The learned class-conditional likelihoods also inherit the coverage, label noise, and policy assumptions of the training data. As a result, improved calibration and reduced shortcut behavior on the tested benchmarks should not be interpreted as a guarantee of robustness to all distribution shifts, adaptive jailbreaks, or future safety taxonomies.

\algo also changes the computational profile of guard inference. A direct verdict-token guard typically requires one forward pass to score the label token. In contrast, \algo estimates two class-conditional reconstruction scores and averages across sampled masks; in the default evaluation this uses $K=3$ sampled masks per label. This additional parallel computation is useful for calibration and token-level evidence, but it can be costly in high-throughput moderation pipelines. Finally, the prompt-rewriting experiment is intended as a proof of concept. 

\subsection{Broader Impact and Ethics}
\label{app:impact-ethics}

The intended positive impact of this work is to improve the reliability of safety moderation systems. Better calibrated guard models can support more transparent refusals, reduce overconfident mistakes, and help distinguish harmful content from benign prompts that contain superficially unsafe terms. The prompt-rewriting pipeline may also help systems offer safer alternatives rather than only blocking user requests.

The same capabilities carry risks. More informative guard scores and token-level localization could be misused to probe a moderation system, optimize around safety filters, or identify which fragments trigger enforcement. A guard model can also be used for overbroad filtering, censorship, or surveillance if deployed under inappropriate policies. Further, a guard model can be used adversarially to generate jailbreak attacks. False positives may block legitimate speech, while false negatives may allow harmful content through. These risks motivate treating \algo as one component of a larger safety stack, with policy-specific validation, monitoring after deployment, and human review for high-impact decisions. Our experiments use existing public safety datasets and released model checkpoints; we do not collect new human-subject data.

\subsection{Compute and Hardware}
\label{app:compute}

We run the fine-tuning for \algo from LLaDA-8B-Instruct, and all experiments on a single system with $1\times$ NVIDIA H100 NVL, 30 Intel Xeon Platinum 8562Y+ cores, and 96 GB RAM. The wall-clock time for a full training run is 6 hours and 45 minutes. The full evaluation time, including baselines, is estimated at 11 hours.

%%%%%%%%%%%%%%%%%%%%%%%%%%%%%%%%%%%%%%%%%%%%%%%%%%%%%%%%%%%%

\newpage

\end{document}